\documentclass{article}

\usepackage{microtype}
\usepackage{graphicx}
\usepackage{subcaption}
\usepackage{booktabs} 
\usepackage{makecell} 

\usepackage{hyperref}

\usepackage[preprint]{icml2026}

\usepackage{amsmath}
\usepackage{amssymb}
\usepackage{mathtools}
\usepackage{amsthm}

\usepackage[capitalize,noabbrev]{cleveref}

\usepackage{microtype}  
\usepackage{graphicx}   
\usepackage{subcaption} 
\usepackage{booktabs}   
\usepackage{multirow} 
\usepackage{siunitx}    

\usepackage{amsmath,amssymb,mathtools} 
\usepackage{bm} 

\usepackage{hyperref} 

\usepackage[textsize=tiny]{todonotes} 

\newcommand{\diag}{\operatorname{diag}}

\newcommand{\vDelta}{\bm{\delta}}

\newcommand{\vEps}{\bm{\varepsilon}}

\newcommand{\vSigma}{\bm{\sigma}}

\newcommand{\mD}{\bm{D}}

\newcommand{\mM}{\bm{M}}

\newcommand{\mR}{\bm{R}}

\newcommand{\mU}{\bm{U}}
\newcommand{\mV}{\bm{V}}

\newcommand{\mSigma}{\bm{\Sigma}}

\newcommand{\sUnit}{\mathcal{U}}   
\newcommand{\sCrys}{\mathcal{X}}   
\newcommand{\sSpec}{\mathcal{Z}}   
\newcommand{\sReal}{\mathbb{R}}    
\newcommand{\sInt}{\mathbb{Z}}     

\newcommand{\nSpec}{z}             
\newcommand{\dModel}{d_{\text{model}}} 
\newcommand{\sLR}{\lambda}       

\newcommand{\vFrac}{\bm{f}}        
\newcommand{\vInt}{\bm{k}}         
\newcommand{\mLatt}{\bm{L}}        
\newcommand{\mSymm}{\bm{S}}        
\newcommand{\vEnc}{\bm{\mu}}       
\newcommand{\vNois}{\bm{\eta}}     
\newcommand{\diffStd}{\bm{\tau}}  

\newcommand{\sWave}{\mathcal{J}}      
\newcommand{\vWave}{\bm{j}}           
\newcommand{\bWave}{\varphi}          
\newcommand{\jMax}{j_{\mathrm{max}}}  

\newcommand{\vFour}{\bm{\alpha}}      
\newcommand{\mFour}{\bm{A}}           

\newcommand{\lVae}{\mathcal{L}_{\text{VAE}}}
\newcommand{\lFour}{\mathcal{L}_{\text{four}}}
\newcommand{\lLat}{\mathcal{L}_{\text{lat}}}
\newcommand{\lMu}{\mathcal{L}_{\mu}}
\newcommand{\lElt}{\mathcal{L}_{\text{CE}}}
\newcommand{\lRec}{\mathcal{L}_{\text{rec}}}
\newcommand{\lEta}{\mathcal{L}_{\eta}}

\DeclareSIUnit{\angstrom}{\text{\AA}}
\DeclareSIUnit{\electronvolt}{\text{eV}}
\DeclareSIUnit{\evperatom}{\text{eV/atom}}

\icmltitlerunning{Fourier-Space Transformers for Latent Diffusion in Crystallography}

\begin{document}

\twocolumn[
\icmltitle{Fourier Transformers for Latent Crystallographic Diffusion and Generative Modeling}


\begin{icmlauthorlist}
  \icmlauthor{Jed A. Duersch}{cril,uccs}
  \icmlauthor{Elohan Veillon}{cril,uccs}
  \icmlauthor{Astrid Klipfel}{cril}
  \icmlauthor{Adlane Sayede}{uccs}
    \icmlauthor{Zied Bouraoui}{cril}
\end{icmlauthorlist}

\icmlaffiliation{cril}{CRIL UMR 8188, Universit\'e d'Artois, CNRS, France}
\icmlaffiliation{uccs}{UCCS UMR 8181, Universit\'e d'Artois, France}

\icmlcorrespondingauthor{Jed A. Duersch}{jalma.duersch@univ-artois.fr}

\icmlkeywords{Crystallography, Diffusion Models, Fourier Representation, Generative Modeling}

\vskip 0.3in
]



\printAffiliationsAndNotice{}  

\begin{abstract}
The discovery of new crystalline materials calls for generative models that handle periodic boundary conditions, crystallographic symmetries, and physical constraints, while scaling to large and structurally diverse unit cells. We propose a reciprocal-space generative pipeline that represents crystals through a truncated Fourier transform of the species-resolved unit-cell density, rather than modeling atomic coordinates directly. This representation is periodicity-native, admits simple algebraic actions of space-group symmetries, and naturally supports variable atomic multiplicities during generation, addressing a common limitation of particle-based approaches. Using only nine Fourier basis functions per spatial dimension, our approach reconstructs unit cells containing up to 108 atoms per chemical species. We instantiate this pipeline with a transformer variational autoencoder over complex-valued Fourier coefficients, and a latent diffusion model that generates in the compressed latent space. We evaluate reconstruction and latent diffusion on the LeMaterial benchmark and compare unconditional generation against coordinate-based baselines in the small-cell regime ($\leq 16$ atoms per unit cell).
\end{abstract}


\section{Introduction}
The discovery of new crystalline materials underpins progress in energy storage, catalysis, and ionic transport.
Useful phenomena may arise in compositionally rich and structurally complex unit cells containing many atoms and multiple chemical species \cite{Mroz2022}.
Yet even for small unit cells, the size of the admissible design space scales rapidly with atom count and compositional diversity.
Exhaustively exploring such high-dimensional design spaces using density functional theory (DFT) \citep{DFT1-hohenberg-kohn,DFT2-kohn-sham} is computationally costly, motivating the development of generative models that can propose physically plausible candidate structures at scale.

Recent generative models (e.g. \cite{mattergen,ADiT,symmcd}) can learn distributions over known crystals and propose new candidates for downstream screening, but extending these approaches to realistic crystallographic regimes remains challenging. Crystals are inherently periodic, with geometry defined on a three-dimensional torus and constrained by crystallographic symmetries \cite{gemsdiff,diffcsp++}, while many practically relevant materials exhibit variable atomic multiplicities and stoichiometries. Coordinate- and particle-based generators handle periodicity and symmetry through geometric conventions and equivariant architectures, but often incur substantial modeling overhead and, in many diffusion formulations, implicitly assume a fixed number of atoms. This limits support for generative moves that change multiplicity and complicates stable behavior across symmetry subgroups when composition changes.

In this work we take a different view: we represent each crystal through the truncated Fourier transform of a species-resolved unit-cell density, and perform learning and generation directly in reciprocal space. This representation is periodicity-native by construction and provides a compact, fixed-shape description that scales with unit-cell complexity. Importantly, crystallographic symmetries act on Fourier coefficients via simple algebraic transformations (permutations of wave vectors and phase shifts), turning symmetry structure into constraints on coefficients rather than geometric operations on particles. Because atomic multiplicity is encoded in the density, the representation also naturally supports variable numbers of atoms per species.

Building on this representation, we propose a two-stage reciprocal-space generative pipeline. We first train a transformer-based variational autoencoder (VAE) operating on complex-valued Fourier coefficients to learn a compact latent code that reconstructs lattice parameters, species information, and truncated Fourier content. We then train a latent diffusion model to generate in this latent space, enabling unconditional sampling followed by decoding and coordinate recovery. Our main contributions are:

\noindent\textbf{(i) Reciprocal-space crystal representation and recovery:} we introduce a truncated Fourier representation of species-resolved unit-cell density together with a practical recovery procedure that maps Fourier coefficients back to atomic configurations.

\noindent\textbf{(ii) Complex-valued transformer VAE with latent diffusion generation:} we built a transformer VAE  for complex Fourier tokens and a latent diffusion model that performs generative sampling in the resulting
compressed space.

\noindent\textbf{(iii) Scalability and multiplicity support.}
With only nine Fourier basis functions per spatial dimension, our representation supports exact recovery for unit cells containing up to 108 atoms per chemical species and enables encoding and denoising of higher-count structures.
In unconditional generation, latent diffusion amplifies the low-count skew present in the meta-stable dataset used for fine-tuning; accordingly, we focus quantitative generation comparisons on the \emph{small-cell regime} ($\leq 16$ atoms per unit cell).

Our experiments validate recoverability from the truncated Fourier representation, stable VAE compression, and robust reverse diffusion in the signal-dominant regime.
Unconditional sampling yields crystals with variable atom counts and symmetry-consistent fractional structure.
These results establish the feasibility of reciprocal-space latent diffusion.


\section{Related Work}
\label{sec:related}

\noindent\textbf{Crystal representations for generative modeling}
A foundational choice in crystal generative modeling is the representation of atomic structure, which largely determines scalability and physical fidelity. Early approaches discretize unit cells into 3D voxel grids and model atomic or electron density as images using convolutional VAEs and GANs \cite{iMatGen,hoffmann-et-al-voxel-vae,court-et-al-voxel-vae,ZeoGAN,CCDCGAN,VAE,GAN}, but these representations are memory-intensive and scale poorly with increasing unit-cell complexity. A second family of methods uses invertible, CIF- or POSCAR-inspired feature tensors to enable valid reconstruction and property conditioning \cite{FTCP,CVAE,wycryst}, at the cost of entangling composition, symmetry, and geometry in ways that can limit robustness under extrapolation. In these settings, periodicity, symmetry, and atom-count constraints are typically handled through explicit design choices (e.g., fixed grids, constrained parameterizations, or post-processing) rather than being made intrinsic to the primary modeling variables. 
Our work follows the density-field perspective (species-resolved unit-cell densities) but moves the primary modeling domain to reciprocal space via a truncated Fourier basis.

\noindent\textbf{Diffusion models for crystal generation.}
Diffusion-based models have become a dominant paradigm for crystal generation due to their strong validity and stability. Most existing approaches operate directly in atomic coordinate space, diffusing positions, lattices, and sometimes atom types under geometric constraints \cite{cdvae,symat,diffcsp,mattergen}, with subsequent work improving energetic fidelity and controllability through refined priors and conditional generation \cite{dp-cdvae,con-cdvae,cond-cdvae,diffcrysgen}. However, coordinate-space diffusion entails substantial geometric and combinatorial complexity, requiring explicit treatment of periodic boundaries, symmetry multiplicity, and typically assuming a fixed number of atoms. Latent-space diffusion can alleviate these issues by reducing dimensionality and accelerating sampling, but only if the latent representation has already factored out periodicity- and symmetry-induced redundancies.

\noindent\textbf{Symmetry-aware generative modeling}
A substantial body of work incorporates crystallographic symmetry explicitly into generative models. Wyckoff-position encodings and space-group constraints can improve validity (and sometimes sample efficiency) by restricting generation to symmetry-consistent degrees of freedom \cite{wycryst,smld,diffcsp++,symmcd,wyckoffdiff}, but often trade generative flexibility for explicit control, require pre-specification of symmetry classes, or increase architectural and training complexity, especially when exploring under-characterized regions of crystal space or transitions between symmetry subgroups.

\noindent\textbf{Transformer-based crystal generators}
Transformers are a natural backbone for crystal generation due to their ability to model long-range interactions and heterogeneous tokens. Autoregressive models such as CrystalFormer and WyFormer employ symmetry-constrained tokenizations informed by Wyckoff positions \citep{CrystalFormer,wyformer}, while more recent diffusion-based transformer systems introduce hybrid discrete--continuous diffusion and property-guided generation pipelines \cite{CDI_CDS,Uni-MDM,ADiT,CrystalDiT}. Despite these advances, transformer-based generators largely inherit the assumptions of their underlying encodings, motivating approaches that address scalability, symmetry, and compositional flexibility through the representation itself (and its latent compression), rather than relying only on architectural constraints.

\noindent\textbf{Alternative paradigms}
Several alternative paradigms, including large language model--based encodings and Generative Flow Networks, have been explored \cite{LLaMA-2-fine-tuned,SLICES,GFlowNets,Crystal-GFN,CHGFlowNets}. These directions are complementary, but they typically focus on discrete interfaces, conditioning, or combinatorial search and therefore do not by themselves remove the representational redundancy induced by periodicity and space-group structure that motivates our reciprocal-space approach.


\section{Overview and Crystal representation}
We propose a reciprocal-space generative pipeline for crystalline materials based on a truncated Fourier representation of the species-resolved unit-cell density. Instead of modeling atomic coordinates directly, we encode each crystal as (i) a rotation-invariant parametrization of the lattice and (ii) complex Fourier coefficients on a fixed reciprocal grid. This choice makes periodic boundary conditions exact by construction and exposes crystallographic symmetries as simple algebraic structure in the modeled variables.

Our pipeline has two stages. First, a \textbf{complex-valued transformer VAE} compresses the Fourier representation into a structured latent code and reconstructs lattice parameters, species information, and truncated Fourier coefficients. Second, a \textbf{latent diffusion model} generates these latent codes, enabling unconditional sampling followed by decoding and coordinate recovery. Because the zero-frequency Fourier coefficient encodes per-species multiplicity, the generator supports variable numbers of atoms per species without requiring a fixed-size point set during sampling. 

\subsection{Representation}
A crystal is defined by a periodic arrangement of atoms in $\sReal^3$.
For our purposes, it is convenient to represent this structure by a unit cell whose contents are partitioned by atomic number.
Let $\sSpec \subset \sInt$ denote the set of atomic numbers present in the material.
For each species $\nSpec \in \sSpec$, we define the corresponding subset of the unit cell by
$\sUnit_{\nSpec}
=
\left\{
\vFrac^{(\nSpec)}_a \in [0,1)^3
\;\middle|\;
a = 1, \dots, n_{\nSpec}
\right\},$
where $n_{\nSpec}$ is the number of atoms of species $\nSpec$,
and fractional coordinates are taken in the half-open cube $[0,1)^3$ to avoid boundary duplication.
Repeating the unit cell over the integer lattice and embedding it in physical space using a lattice matrix $\mLatt \in \sReal^{3 \times 3}$ yields the atomic positions
\[
\sCrys_{\nSpec}
=
\left\{
\mLatt \bigl( \vFrac + \vInt \bigr)
\;\middle|\;
\vFrac \in \sUnit_{\nSpec},\;
\vInt \in \sInt^3
\right\}.
\]
The complete crystal configuration is the disjoint union over species present
$\sCrys = \bigsqcup_{\nSpec \in \sSpec} \sCrys_{\nSpec}$.


Material properties are invariant under global rotations of the crystallographic axes. We therefore adopt a rotation-invariant and well-conditioned parametrization of lattice geometry using the matrix logarithm of its symmetric polar factor \cite{diffcsp}. This yields an additive, rotation-invariant encoding of lattice shape that is stable across a wide range of length scales and well suited for learning and optimization. Full details are provided in Appendix~\ref{app:lattice_param}.

\subsection{Truncated Fourier Transform of Atomic Density}
One of the primary difficulties with diffusion models operating directly on atomic coordinates is the coherent handling of periodic boundary conditions. For example, nearest-neighbor computations require explicit replication across all $27$ translations in $\{-1,0,1\}^3$, complicating graph-based or locality-sensitive
architectures. Moreover, coordinate-wise diffusion does not naturally permit changes in atomic composition during the generative process.
We instead construct a variational autoencoder that operates on the Fourier representation of species-resolved atomic density functions. This representation
naturally enforces periodicity while allowing explicit control over the
composition of each atomic species.
Let $\sWave \subset \sInt^3$ denote a truncated set of Fourier wave vectors,
\[
\sWave = \left\{ \vWave \in \sInt^3 \;\middle|\; \|\vWave\|_\infty \le \jMax \right\}.
\]
Each $\vWave \in \sWave$ defines a Fourier basis function on the unit cube,
$\bWave_{\vWave}(\vFrac) = \exp\!\left(2\pi i\, \vWave^\top \vFrac \right).$
Representing each atom as a Dirac delta distribution, the Fourier coefficients
of the atomic density for species $\nSpec$ are given by
\[
\vFour_{\vWave}^{(\nSpec)}
= \sum_{\vFrac \in \sUnit_{\nSpec}}
\exp\!\left(-2\pi i\, \vWave^\top \vFrac \right),
\qquad \vWave \in \sWave.
\]
For each species, these coefficients combine to yield a vector
$\vFour^{(\nSpec)} = [\vFour_{\vWave}^{(\nSpec)}]_{\vWave \in \sWave}$.
The vectors are then assembled into a matrix representation of the full crystal,
\[
\mFour =
\begin{bmatrix}
\vFour^{(\nSpec_1)} & \vFour^{(\nSpec_2)} & \cdots & \vFour^{(\nSpec_m)}
\end{bmatrix}.
\]
Since the number of atoms in each species is contained in the $\vFour_0$ coefficients,
we easily know how many fractional coordinates to reconstruct from the Fourier coefficients.

\noindent\textbf{Symmetries and algebraic structure}
In Fourier space, crystallographic symmetries induce simple algebraic constraints
among reciprocal-space coefficients. Specifically, a symmetry operation consisting of a lattice-preserving rotation $\mM$ and a fractional translation $\vDelta$ acts by permuting wave vectors and introducing phase shifts, yielding the relation
\[
\vFour_{\vWave}^{(\nSpec)} =
\exp\!\left(-2\pi i\, \vWave^\top \vDelta\right)\,
\vFour_{\mM^\top \vWave}^{(\nSpec)}.
\]
These structured relationships provide a natural mechanism for symmetry-aware compression and generation; full details are given in Appendix~\ref{app:symmetry_algebra}.


\section{Variational encoding and reverse diffusion}

\begin{figure}[t]
\centering
\begin{subfigure}[t]{\columnwidth}
  \centering
  \includegraphics[width=0.75\columnwidth]{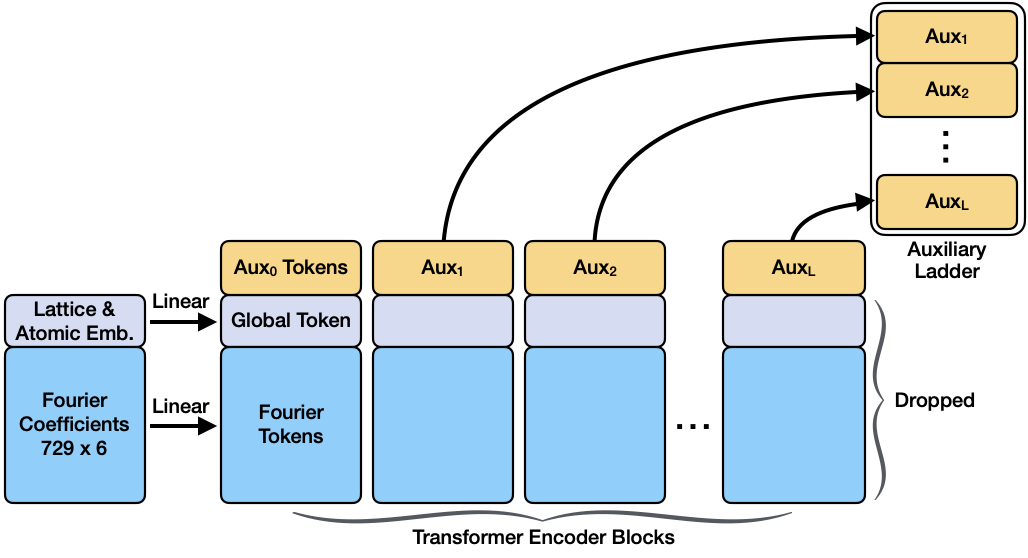}
  \caption{Encoder: $L$ complex transformer blocks  process auxiliary, global,
and Fourier tokens. Auxiliary tokens are extracted at each depth and concatenated
to form a depth-aligned auxiliary ladder.}
  \label{fig:aux_fourier_encoder}
\end{subfigure}
\vspace{0.6ex}
\begin{subfigure}[t]{\columnwidth}
  \centering
  \includegraphics[width=0.75\columnwidth]{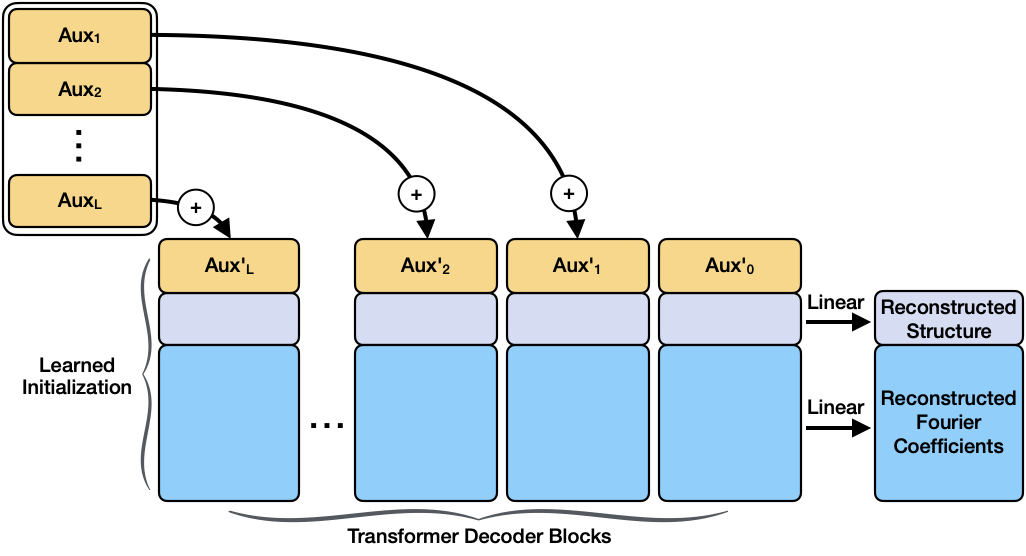}
  \caption{Decoder: tokens are reinitialized at the bottleneck. Depth-matched ladder slices are added to auxiliary tokens before each block to condition reconstruction.}
  \label{fig:aux_fourier_decoder}
\end{subfigure}
\caption{\textbf{Aux--Global--Fourier encoder--decoder with auxiliary ladder.}
The model separates tokens into a small auxiliary block, a single global structure
token, and a large set of Fourier tokens. 
\textbf{(a)} The encoder distills information into a depth-aligned auxiliary ladder.
\textbf{(b)} The decoder reconstructs structure and Fourier coefficients by injecting
auxiliary ladder slices at matching depths.}
\label{fig:aux_fourier_unet_perceiver}
\end{figure}

Our architecture combines a Perceiver-style bottleneck \citep{jaegle2021perceiver} with U-Net-style depth-aligned conditioning \citep{ronneberger2015u} for reciprocal-space crystal representations. A small auxiliary token block interfaces with the large set of Fourier tokens: the encoder distills the full sequence into a depth-aligned auxiliary ladder, and the decoder reconstructs from learned token initializations by injecting the corresponding ladder slices at each layer. This bottlenecked, depth-matched conditioning stabilizes reconstruction under strong compression.
Figure~\ref{fig:aux_fourier_unet_perceiver} summarizes the encoder–decoder. The crystal input is embedded as a heterogeneous token sequence in a shared complex feature space (Appendix~\ref{app:tokenization}).

\noindent\textbf{Symmetry-aware latent compression}
%
In reciprocal space, crystallographic symmetries relate Fourier coefficients associated with rotated wave vectors, while translational symmetries appear as phase shifts.
We do not explicitly enforce space-group constraints during training.
Instead, we expose symmetry-relevant structure through
(i) a fixed reciprocal-space grid in which each retained wave vector $\vWave$ is represented as a token, so that rotational symmetries act as permutations of tokens, and
(ii) a Fourier-aware complex rotary positional encoding tied directly to $\vWave$ (Appendix~\ref{app:complex_transformer}), which preserves relative phase structure.
The auxiliary bottleneck encourages the model to compress global regularities, including symmetry-consistent correlations across wave vectors and species, into a compact latent code.
We first train a VAE to reconstruct lattice parameters and Fourier coefficients, and subsequently train a latent diffusion model to sample these codes for unconditional generation.


%
%
\subsection{Encoder--decoder architecture with auxiliary ladder}



Each encoder and decoder layer is a complex-valued transformer block (RMSNorm, multi-head self-attention, gated MLP), implemented in complex space while remaining compatible with optimized real-valued attention kernels. Complex operations preserve phase throughout attention and residual streams, enabling rotary positional encodings tied to reciprocal-space wave vectors.
In the baseline, the gated MLP modulates real and imaginary parts independently; Section~\ref{sec:ablation} suggests a phase-preserving variant based on modulus gating can improve optimization efficiency without disrupting phase coherence. We use a Fourier-aware complex rotational encoding for positions; full details are in Appendix~\ref{app:complex_transformer}. The VAE is a symmetric complex encoder–decoder that distills information into a compact auxiliary code and re-injects it during reconstruction (Fig.~\ref{fig:aux_fourier_unet_perceiver}).

\noindent\textbf{Encoder and auxiliary ladder construction}
The encoder consists of $L$ stacked complex transformer blocks operating on the
full token sequence, including auxiliary tokens, a single global structure token,
and a large set of Fourier tokens. While all tokens participate in attention at
each depth, only the auxiliary token block is retained after each transformer
layer.
Concretely, after the $\ell$-th encoder block, the auxiliary token state
$\mathrm{Aux}_\ell \in \mathbb{C}^{n_{\text{aux}}\times \dModel}$ is
extracted and stored. Concatenating these states across depth yields a
depth-aligned auxiliary ladder,
\[
Z = [\mathrm{Aux}_1;\mathrm{Aux}_2;\cdots;\mathrm{Aux}_L],
\]
which serves as the latent representation passed to the variational bottleneck. Fourier and global tokens act as contextual carriers during encoding but are not
explicitly preserved.
To encourage compact latent representations, auxiliary channels are progressively suppressed using a learned channel-wise mask trained under a signal-to-noise regularization scheme. This enables adaptive capacity allocation and permanent pruning of non-informative channels during training. Full details are provided in Appendix~\ref{app:aux_ladder}.

%
\noindent\textbf{Variational bottleneck}
The (masked) auxiliary ladder $\vEnc \in \mathbb{C}^{(L n_{\text{aux}})\times \dModel}$ parameterizes the mean of a diagonal complex Gaussian latent distribution.
We associate each latent dimension with a learned real-valued log-scale $\vSigma$ and, using noise $\vEps$ that is independent in real and imaginary components, sample the latent variable passed to the decoder as
\[
Z = \vEnc + \vEps \odot \exp(\vSigma),
\qquad
\vEps \sim \mathcal{N}(0,I).
\]
\noindent\textbf{Decoder and depth-aligned conditioning}
The decoder mirrors the encoder with $L$ complex transformer blocks but operates
on a reinitialized token sequence. All tokens—auxiliary, global, and Fourier—are
initialized from learned constants. At each decoder depth, a slice of the latent
auxiliary ladder is injected additively into the auxiliary token block in
depth-matched reverse order.
Specifically, before the $i$-th decoder block, the corresponding auxiliary slice
is added only to the auxiliary tokens, while the remaining tokens are left
unchanged. This auxiliary-only injection conditions the decoder on the encoded
latent state without directly copying token content, allowing the decoder to
reconstruct global structure and Fourier coefficients from scratch.
After decoding, lightweight reconstruction heads map the decoded tokens to lattice
parameters, species assignments, and Fourier coefficients.
Details of the reconstruction heads and the multistage coordinate recovery
procedure are provided in Appendix~\ref{app:reconstruction}.

\subsection{Reverse diffusion in the auxiliary latent space}

The role of this model
is to denoise corrupted auxiliary ladders by predicting the injected noise, enabling
iterative refinement in latent space.

\noindent\textbf{Implementation}
The reverse diffuser is implemented as a complex-valued transformer with the same
number of blocks as the encoder or decoder. Its input consists of a learned
initialization tensor whose token dimension matches the size of the auxiliary
ladder, augmented with additional scratch-pad tokens that are positionally encoded
using the same Fourier-aware rotational scheme as the Fourier tokens. This learned
initialization provides a nonzero baseline representation for all tokens.
The noise-corrupted auxiliary ladder is added to the leading auxiliary portion of
this initialization.
This additive conditioning ensures that channels suppressed during encoding retain
representational capacity during denoising and can participate in attention and
intermediate computation.
Several model components are conditioned on the diffusion coordinate using a lightweight quadratic interpolation scheme, enabling time-dependent modulation without introducing auxiliary networks. Full details are provided in Appendix~\ref{app:diffusion}.

\noindent\textbf{Selector-aware denoising}
To maintain consistency with the variational encoder, the reverse diffuser reuses the binary channel selector learned during encoding, applying denoising only to latent channels retained after auxiliary compression.
This avoids imposing artificial constraints on the diffuser to reproduce exact zeros in pruned channels, which never encounter injected noise.
After $L$ transformer blocks, the diffuser outputs a noise estimate corresponding to the auxiliary ladder tokens, and training proceeds by comparing this prediction to the known injected noise under a standard diffusion objective.


\section{Training Methodology}
\label{sec:training}
Training consists of two stages with coupled objectives: (i) training a variational autoencoder (VAE) to learn a compressible latent representation and reconstruct lattice/species/Fourier content, and (ii) training a reverse diffusion model to denoise corrupted latent codes for generative sampling. To keep optimization stable across heterogeneous objectives, we normalize losses so that their effective gradient scales remain comparable: reconstruction terms are combined through a scale-invariant aggregation (Appendix~\ref{app:recon_aggregation}), and diffusion noise-prediction errors are normalized by per-channel noise scales (Appendix~\ref{app:diffusion_noise}).
The encoder is shared between the VAE and diffusion stages. During VAE training, the encoder produces an auxiliary-ladder latent representation that is regularized to encourage compact, suppressible channels while maintaining reconstruction fidelity. During diffusion training, the encoder output serves as the target signal for a noise-conditioned denoising process operating directly in the auxiliary latent space.

\subsection{Variational Auto-Encoder Training Objectives}
\label{sec:vae_training}

The decoder predicts three components: lattice coefficients, species-slot assignments, and complex Fourier coefficients.

\noindent\textbf{Species prediction}
Atomic numbers are predicted independently for each of up to six species slots using a cross-entropy loss over logits. An additional ``empty'' class accounts for unused slots.

\noindent\textbf{Lattice reconstruction}
Lattice geometry is represented by six real-valued coefficients. Reconstruction error is measured using mean squared error (MSE),
$\lLat = \mathrm{MSE}(\hat s, s)$.

\noindent\textbf{Fourier reconstruction}
For each species slot, complex Fourier coefficients are reconstructed using the squared complex modulus, equivalently the sum of squared errors over real and imaginary components.
\begin{equation}
\lFour =
\frac{1}{6 |\sWave|}\sum_{z=0}^{5}
\left\lVert \hat{\vFour}^{(z)} - \vFour^{(z)} \right\rVert_2^2,
\end{equation}
\noindent\textbf{Scale-Invariant Reconstruction Aggregation}
Lattice and Fourier reconstruction losses are combined into a single reconstruction term $\lRec = \sqrt{\lLat + \lFour}$,
forming a scale-invariant aggregation that maintains balanced gradient contributions across reconstruction terms throughout training. Details and motivation are provided in Appendix~\ref{app:recon_aggregation}.

\noindent\textbf{Latent Signal-to-Noise Regularization}
Latent regularization is implemented using a signal-to-noise–based penalty that
encourages compact, suppressible latent channels while avoiding the vanishing
gradients associated with Gaussian KL regularization.
Formal definitions and optimization rationale are given in
Appendix~\ref{app:snr_latent}.

\noindent\textbf{Full VAE Objective}
The training objective is
\begin{equation}
\lVae =
\mathbb{E}_{\text{batch}}
\left[
\lambda_z \,\lElt(\hat z, z)
+ \lRec
+ \lambda_\mu \,\lMu
\right],
\end{equation}
where $\lambda_z$ controls the contribution of species classification and $\lambda_\mu$ controls the strength of latent SNR regularization. 
The reconstruction term $\lRec$ has unit weight as a reference, with other terms weighted relative to it.

\subsection{Reverse Diffusion Training}
\label{sec:reverse_diffusion_training}

\noindent\textbf{Latent Noise Model and Scale Estimation}
Reverse diffusion operates on noise-corrupted latent representations using a
per-channel noise model with learned characteristic scales.
The full noise distribution, moment matching, and scale estimation procedure are
described in Appendix~\ref{app:diffusion_noise}.

\noindent\textbf{Diffusion Schedule and Information-Matched Mixing}
Latent variables are corrupted using a diffusion schedule designed to be
approximately linear in expected information content, enabling stable training
across signal- and noise-dominated regimes.
The full schedule construction and parameterization are given in
Appendix~\ref{app:diffusion_schedule}.

\noindent\textbf{Reverse Diffusion Objective}
Given $(\vEnc_\phi, \phi)$, the reverse diffuser predicts the sampled noise $\vNois$. The training loss is the root-mean-squared error of the noise prediction, normalized by the decoder noise scale,
\begin{equation}
\lEta = \mathbb{E}
\left[ \left\lVert (\hat{\vNois} - \vNois)/\tau \right\rVert_2^2 \right]^{1/2}.
\end{equation}
While signal reconstruction $\hat{\vEnc} = \frac{\vEnc_\phi - s_\phi \hat{\vNois}}{c_\phi}$ is monitored during training, the signal error is not directly optimized.

\noindent\textbf{Rationale for Noise Prediction}
Empirically, the most challenging regime for the reverse diffuser is the signal-dominant mixture, where $s_\phi$ is small and the noise contribution is weakly identifiable from $\vEnc_\phi$. In this regime, both noise prediction error and signal reconstruction error are largest. Optimizing a direct signal-reconstruction objective was found to be less stable and yielded inferior performance.
The noise-prediction objective provides a well-conditioned regression target and naturally concentrates gradient pressure on the empirically limiting regime, without requiring additional schedule-dependent weighting. Combined with the information-matched diffusion schedule, this yields a robust and efficient reverse diffusion training procedure.


\section{Experiments}
\label{sec:experiments}

\subsection{Experimental protocol}
\label{sec:protocol}

\noindent\textbf{LeMaterial}
provides a large, harmonized collection of crystalline structures aggregated from major repositories (Materials Project, Alexandria, OQMD), designed for large-scale ML in materials science \citep{lematerial_2025}. We use the pre-filtered release containing approximately $2.8$ million structures, most of which are DFT-relaxed local minima and therefore not guaranteed to be experimentally synthesizable. Many practically relevant candidates are \emph{meta-stable}, i.e., near the formation-energy Pareto frontier for a given stoichiometry rather than at the global minimum. Filtering to meta-stable crystals yields $506{,}853$ structures. Within this subset, $74.2\%$ have at most $16$ atoms per unit cell (the \emph{small-cell regime}). We reserve $2048$ meta-stable structures for testing using a stratified split over unit-cell complexity, including $512$ small-cell examples; all remaining meta-stable structures are used for fine-tuning.


\noindent\textbf{Preprocessing and recoverability}
All crystals are preprocessed by converting species-resolved unit-cell densities to a truncated Fourier representation and snapping fractional coordinates to a rational grid to enable reliable inversion. We consider two regimes: (i) $7$ basis functions per dimension ($7^3=343$ modes) with snapping to a $1/24$ grid, and (ii) $9$ basis functions per dimension ($9^3=729$ modes) with snapping to a $1/48$ grid. On the full pre-filtered dataset, regime (i) flags $101{,}440$ structures as unrecoverable, whereas regime (ii) flags only $27{,}005$. Restricting to meta-stable subset, the corresponding counts are $40{,}793$ versus $10{,}902$. We therefore use $\text{bpd}=9$ in experiments.

\begin{table}[t]
    \centering
    \footnotesize
    \caption{\textbf{Head dimension sweep} (vary $D_{\mathrm{head}}$; baseline $N_{\mathrm{head}}=12$, $N_{\mathrm{layer}}=8$).
    Width changes via $d_{\mathrm{model}}=N_{\mathrm{head}}D_{\mathrm{head}}$.
    Best (lowest) values per column are bolded.}
    \label{tab:dhead}
    \begin{tabular}{c||c|c|c|c|c}
    \toprule
    $D_{head}$ & $\lVae$ & $\lFour$ & $\lLat$ & $\lMu$ & $\lElt$ \\
    \midrule
    36 & 0.264 & 0.228 & 0.070 & 0.038 & 0.003 \\
    \textbf{48} & \textbf{0.221} & 0.209 & \textbf{0.034} & 0.043 & 0.001 \\
    60 & 0.232 & \textbf{0.193} & 0.072 & \textbf{0.042} & 0.004 \\
    \bottomrule
    \end{tabular}
\end{table}
\begin{table}[t]
    \centering
    \footnotesize
    \caption{\textbf{Head count sweep} (vary $N_{\mathrm{head}}$; baseline $D_{\mathrm{head}}=48$, $N_{\mathrm{layer}}=8$).
    Width changes via $d_{\mathrm{model}}=N_{\mathrm{head}}D_{\mathrm{head}}$.
    Best (lowest) values per column are bolded.}
    \label{tab:nhead}
    \begin{tabular}{c||c|c|c|c|c}
    \toprule
    $N_{head}$ & $\lVae$ & $\lFour$ & $\lLat$ & $\lMu$ & $\lElt$ \\
    \midrule
    10 & \textbf{0.216} & \textbf{0.186} & 0.061 & 0.045 & 0.001 \\
    11 & 0.229 & 0.195 & 0.069 & 0.039 & 0.001 \\
    12 & 0.221 & 0.209 & \textbf{0.034} & 0.043 & 0.001 \\
    14 & 0.269 & 0.236 & 0.072 & \textbf{0.033} & 0.001 \\
    \bottomrule
    \end{tabular}
\end{table}

\noindent\textbf{Training overview}
%
VAE is trained on the full preprocessed corpus (not only the meta-stable subset) to learn a broad reciprocal-space representation; encoding meta-stability directly in the latent space is non-trivial, so we impose this bias at generation time.
The reverse diffuser is trained in two stages: pretraining on the full dataset to learn a general prior over latents, followed by fine-tuning on the meta-stable training split to steer sampling toward thermodynamically favorable structures.
%
All experiments use a fixed data/model interface with maximum atomic number $z_{\max}=83$ and at most $6$ chemical species per crystal. Unless otherwise noted, Fourier inputs use a cubic truncation with $\text{bpd}=9$ basis functions per dimension and are evaluated on a $\text{gpd}=48$ grid per dimension.
We train with batch size $8$ using Adam ($\beta_1=0.9$, $\beta_2=0.999$, $\epsilon=10^{-8}$).
%
The learning rate schedule consists of a linear warmup of $1000$ steps from
$\texttt{start\_factor}\cdot \sLR$ to $\sLR$, followed by cosine annealing over
$\texttt{anneal\_steps}=325{,}000$ steps down to $\sLR_{\min}=2\times 10^{-5}$,
with base learning rate $\sLR=2\times 10^{-4}$.
All runs are trained for one epoch (approximately $333{,}727$ steps at batch size $8$ for $\text{bpd}=9$).

\noindent\textbf{Baseline architecture}
Our baseline transformer VAE uses head dimension $D_{\mathrm{head}}=48$,
$N_{\mathrm{head}}=12$ attention heads, and $N_{\mathrm{layer}}=8$ transformer blocks in both encoder and decoder.
We parameterize model width as
$d_{\mathrm{model}} = N_{\mathrm{head}} \, D_{\mathrm{head}}$.
Compression is controlled by a target nonzero-element budget factor $C=8/12$
and a cosine schedule over $\texttt{nnz steps}=160\text{k}$ steps (details below).

\subsection{Representation fidelity and recoverability}
\label{sec:recoverability}

The truncated Fourier representation requires selecting a subset of retained wave vectors $\sWave$.
While spherical truncation ($\|\vWave\|_2 \le \jMax$) is analytically attractive for isolated atoms, we find it poorly matched to crystalline configurations, which often exhibit separable structure across coordinate directions.
Across a large corpus of crystals, cubic truncation ($\|\vWave\|_\infty \le \jMax$) consistently yields higher reconstruction success rates than spherical or hybrid truncations, even when the latter retain more basis functions.

Using cubic truncation, reconstruction success varies smoothly with atomic count and unit-cell occupancy.
We observe successful reconstruction of unit cells containing up to $52$ atoms of a single species using $7^3$ retained modes ($\jMax=3$), and up to $108$ atoms using $9^3$ modes ($\jMax=4$).
These trends are consistent with the recoverability statistics reported in Section~\ref{sec:protocol}, and further motivate the use of higher Fourier resolution when targeting structurally complex or high-occupancy unit cells.

\begin{table}[t]
    \centering
    \footnotesize
    \caption{\textbf{Depth sweep} (vary $N_{\mathrm{layer}}$; baseline $D_{\mathrm{head}}=48$, $N_{\mathrm{head}}=12$).
    Best (lowest) values per column are bolded.}
    \label{tab:nlayer}
    \begin{tabular}{c||c|c|c|c|c}
    \toprule
    $N_{layer}$ & $\lVae$ & $\lFour$ & $\lLat$ & $\lMu$ & $\lElt$ \\
    \midrule
    7 & 0.238 & 0.218 & 0.052 & 0.040 & 0.001 \\
    8 & 0.221 & 0.209 & \textbf{0.034} & 0.043 & 0.001 \\
    9 & \textbf{0.218} & \textbf{0.189} & 0.063 & \textbf{0.038} & 0.001 \\
    \bottomrule
    \end{tabular}
\end{table}
\begin{table}[t]
\centering
\footnotesize
\caption{\textbf{Reverse diffusion pretraining performance.}
All values are RMSE.
Noise RMSE is the optimized training objective.
Signal RMSE (diagnostic) is reported only for the signal-dominant bin and is normalized by the decoder noise scale.
Lower is better.}
\label{tab:rd_pretrain}
\begin{tabular}{c||c|c|c}
\toprule
Model
& low-$\phi$ noise
& mean noise
& low-$\phi$ sig. \\
\midrule
Baseline
& 5.063
& 2.734
& \textbf{0.603} \\
$N_{\mathrm{head}}=10$
& \textbf{4.272}
& \textbf{2.400}
& 0.611 \\
\bottomrule
\end{tabular}
\end{table}
\begin{table}[t]
\centering
\footnotesize
\caption{Reverse diffusion fine-tuning on the meta-stable subset. Results  are for a learning rate of $2\times10^{-5}$.
Lower is better.}
\label{tab:rd_finetune}
\begin{tabular}{c||c|c|c}
\toprule
Model
& low-$\phi$ noise
& mean noise
& low-$\phi$ sig. \\
\midrule
Baseline
& 5.419
& 3.018
& 0.633 \\
$N_{\mathrm{head}}=10$
& \textbf{4.674}
& \textbf{2.661}
& \textbf{0.604} \\
\bottomrule
\end{tabular}
\end{table}
\subsection{VAE reconstruction performance}
\label{sec:vae_results}


\newcommand{\tAngle}{80}
\newcommand{\tSymb}{T}
\newcommand{\fSymb}{F}

\begin{table}[ht]
\centering
\footnotesize
\caption{\textbf{Path ablation over five architectural toggles.}
Each row corresponds to an 80k-step VAE training run following a structured
enable/disable path. Reported losses are averages over the final 1200 steps.
Consistently beneficial flips are bold.}
\label{tab:vae_ablation_path}
\setlength{\tabcolsep}{3.5pt} 
\renewcommand{\arraystretch}{1.05}
\begin{tabular}{ccccc||c}
\toprule
  \rotatebox{\tAngle}{\makecell{cyclic-slots}}
& \rotatebox{\tAngle}{\makecell{rms-bias}}
& \rotatebox{\tAngle}{\makecell{head-scale}}
& \rotatebox{\tAngle}{\makecell{mlp-bias}}
& \rotatebox{\tAngle}{\makecell{modulus-gating}}
& \rotatebox{\tAngle}{\makecell{$\lVae$}} \\
\midrule
  \fSymb & \fSymb & \fSymb & \fSymb & \fSymb & 0.390 \\
  \textbf{\tSymb} & \fSymb & \fSymb & \fSymb & \fSymb & \textbf{0.318} \\
  \tSymb & \tSymb & \textbf{\fSymb} & \fSymb & \fSymb & 0.384 \\
  \tSymb & \tSymb & \tSymb & \fSymb & \fSymb & 0.616 \\
  \tSymb & \tSymb & \tSymb & \tSymb & \fSymb & 0.512 \\
  \textbf{\tSymb} & \tSymb & \tSymb & \tSymb & \textbf{\tSymb} & 0.472 \\
  \fSymb & \tSymb & \tSymb & \tSymb & \tSymb & 0.562 \\
  \fSymb & \fSymb & \tSymb & \tSymb & \tSymb & 0.615 \\
  \fSymb & \fSymb & \textbf{\fSymb} & \tSymb & \tSymb & 0.416 \\
  \fSymb & \fSymb & \fSymb & \fSymb & \textbf{\tSymb} & 0.346 \\
\bottomrule
\end{tabular}
\end{table}

We perform coordinate-wise architectural sweeps over attention factorization ($D_{\mathrm{head}}, N_{\mathrm{head}}$), depth ($N_{\mathrm{layer}}$), and auxiliary compression parameters.
Only one hyperparameter is varied per sweep;
Accordingly, the minima across Tables~\ref{tab:dhead}--\ref{tab:nnzsteps} need not correspond to a jointly
optimal configuration.
For each configuration, we report results from seed achieving the lowest total objective $\lVae$ for that configuration. $\lVae$ is the total weighted training objective, decomposed into Fourier reconstruction loss $\lFour$,
latent loss $\lLat$, mean loss $\lMu$, and element/species loss $\lElt$.
Across all configurations, the element/species cross-entropy loss $\lElt$ was consistently low and exhibited little sensitivity to architectural choices; we therefore treat it as non-limiting and do not use it to guide design decisions.
Model width is parameterized as $d_{\mathrm{model}}=N_{\mathrm{head}}D_{\mathrm{head}}$.
The auxiliary ladder uses a cosine schedule on the number of nonzero elements (nnz),
interpolating from a dense representation to a target nnz determined by the $C$
factor over a specified number of steps.

Across sweeps, we observe a consistent tradeoff between Fourier reconstruction fidelity and latent regularization. While the best coordinate-wise configuration is close to the baseline, the head-count
sweep in Table~\ref{tab:nhead} indicates that reducing the number of attention heads
can yield lower total VAE loss $\lVae$.
Depth sweeps (Table~\ref{tab:nlayer}) further suggest that additional layers may be
beneficial, although these effects are not jointly optimized. Motivated by VAE performance, we compare both the baseline and the
$N_{\mathrm{head}}=10$ architecture in subsequent reverse diffusion experiments.


\subsection{Latent diffusion results}
\label{sec:diffusion_results}
We evaluate reverse diffusion performance using the noise reconstruction
root-mean-squared error (RMSE), which is the quantity directly optimized during
training.
During training, we partition results by the diffusion coordinate $\phi$ into five bins
$\phi \in [0.01,0.2], [0.2,0.4], \ldots, [0.8,0.99]$.
Reconstrution is most challening in the signal-dominant bin ($\phi \in [0.01,0.2]$).
We additionally report the signal reconstruction RMSE as a diagnostic in this regime,
which corresponds to one-shot signal reconstruction error, normalized by the decoder noise scale, and indicates whether the reconstructed signal lies within the range of corruption the decoder is trained to interpret.

\noindent\textbf{Pretraining comparison}
Table~\ref{tab:rd_pretrain} compares reverse diffusion pretraining performance between the baseline architecture and the $N_{\mathrm{head}}=10$ variant.
The modified architecture substantially improves noise reconstruction in the signal-dominant regime, while also reducing the mean noise RMSE across the diffusion schedule.
Signal reconstruction quality in the first bin remains comparable between architectures.

\noindent\textbf{Meta-stable fine-tuning}
We fine-tune the reverse diffuser on the meta-stable subset.
Performance is stable across learning rates spanning more than an order of magnitude; Table~\ref{tab:rd_finetune} reports the best values observed for each architecture. As in pretraining, $N_{\mathrm{head}}$=10 architecture consistently achieves lower noise RMSE in the signal-dominant regime as well as lower mean noise RMSE. Signal reconstruction error in the first bin remains below the decoder noise scale, meaning the reconstructed signal is within the decoder’s operating regime. Full bin-wise results for fine-tuning across learning rates are in Appendix~\ref{app:rd_robustness}.
The results show that the modified architecture improves reverse diffusion performance where it is most challenging (when the noise component is small), while maintaining stable behavior across the full diffusion schedule and fine-tuning learning rates.

\noindent\textbf{Unconditional generation}
$252{,}549$ samples yielded $250{,}473$ small-cell structures ($\leq 16$ atoms) and $2{,}076$ medium-cell structures (17--32 atoms), with no larger unit cells observed.
Since $74.2\%$ of the meta-stable training set lies in the small-cell regime, this indicates amplification of dataset skew under unconditional diffusion, while still demonstrating support for variable atomic multiplicities.

\subsection{Ablation study}
\label{sec:ablation}
To isolate the optimization impact of individual architectural components, we
conduct a path-based ablation study over five binary design toggles. Rather than
evaluating all $2^5$ combinations, we follow a structured enable--disable path
that incrementally adds components to a minimal configuration and then removes
them in reverse order, capturing interaction effects while keeping the number of
runs tractable.
All models are trained for 80k steps under identical settings. Performance is
evaluated using the full VAE objective $\lVae$.

Overall, the ablation highlights the sensitivity of optimization to architectural
biases and their interactions with the auxiliary ladder. Several components that
were beneficial in earlier prototypes no longer improve performance in the
current formulation; in particular, per-head output scaling consistently degrades
optimization. In contrast, cyclic species-slot assignment yields the largest and most consistent improvement, while phase-preserving modulus gating provides a smaller but repeatable gain. Other effects appear interaction-dependent, requiring additional seeds to resolve.

\section{Conclusion}
We presented a reciprocal-space generative pipeline for crystalline materials based on a truncated Fourier representation of species-resolved unit-cell density. This encoding makes periodicity exact, exposes crystallographic symmetries as simple algebraic structure in reciprocal space, and avoids fixed-size point-set assumptions to enable variable atomic multiplicities during generation.
We instantiate this representation with a complex-valued transformer VAE that compresses Fourier content into a structured auxiliary latent ladder, together with a latent reverse diffusion model that operates directly in this compressed space.

Empirically, we show that modest Fourier resolution supports exact coordinate recovery on rational grids for structurally complex, multi-species unit cells, and that reverse diffusion remains stable in the signal-dominant regime under appropriate architectural choices. While unconditional diffusion amplifies the small-cell regime present in the meta-stable training distribution, the encoder and diffuser remain capable of processing higher atom-count structures, indicating that conditioned or guided diffusion is a promising direction for extending generative coverage. Overall, these results position reciprocal-space modeling as a scalable alternative to coordinate-based crystal generators.

\noindent\textbf{Impacts Statement}
This work advances generative modeling for crystalline materials by introducing a reciprocal-space formulation that natively handles periodicity and supports scalable latent diffusion. By enabling efficient modeling of structurally complex crystals, the proposed approach may accelerate computational materials discovery and screening workflows. The work is intended for scientific research applications and does not raise foreseeable negative societal impacts.

\section*{Acknowledgment}
This work was supported by ANR-22-EXES-0009 MAIA and CNRS MITI interdisciplinary programs (PRIME AIM-GPT), and was granted access to the HPC resources of IDRIS under the allocation made by GENCI.

\bibliography{references}

@article{DFT1-hohenberg-kohn,
  title={Inhomogeneous electron gas},
  author={Hohenberg, Pierre and Kohn, Walter},
  journal={Physical review},
  volume={136},
  number={3B},
  pages={B864},
  year={1964},
  publisher={APS}
}

@article{DFT2-kohn-sham,
  title={Self-consistent equations including exchange and correlation effects},
  author={Kohn, Walter and Sham, Lu Jeu},
  journal={Physical review},
  volume={140},
  number={4A},
  pages={A1133},
  year={1965},
  publisher={APS}
}

@article{pbe_functional,
  title={Generalized gradient approximation made simple},
  author={Perdew, John P and Burke, Kieron and Ernzerhof, Matthias},
  journal={Physical review letters},
  volume={77},
  number={18},
  pages={3865},
  year={1996},
  publisher={APS}
}

@article{VAE,
  title={Auto-encoding variational bayes},
  author={Kingma, Diederik P},
  journal={arXiv preprint arXiv:1312.6114},
  year={2013}
}

@article{GAN,
  title={Generative adversarial nets},
  author={Goodfellow, Ian and Pouget-Abadie, Jean and Mirza, Mehdi and Xu, Bing and Warde-Farley, David and Ozair, Sherjil and Courville, Aaron and Bengio, Yoshua},
  journal={NeurIPS},
  volume={27},
  year={2014}
}

@article{GFlowNets,
  title={Flow network based generative models for non-iterative diverse candidate generation},
  author={Bengio, Emmanuel and Jain, Moksh and Korablyov, Maksym and Precup, Doina and Bengio, Yoshua},
  journal={Advances in Neural Information Processing Systems},
  volume={34},
  pages={27381--27394},
  year={2021}
}

@article{hoffmann-et-al-voxel-vae,
  title={Data-driven approach to encoding and decoding 3-d crystal structures},
  author={Hoffmann, Jordan and Maestrati, Louis and Sawada, Yoshihide and Tang, Jian and Sellier, Jean Michel and Bengio, Yoshua},
  journal={arXiv:1909.00949},
  year={2019}
}

@article{iMatGen,
  title={Inverse design of solid-state materials via a continuous representation},
  author={Noh, Juhwan and Kim, Jaehoon and Stein, Helge S and Sanchez-Lengeling, Benjamin and Gregoire, John M and Aspuru-Guzik, Alan and Jung, Yousung},
  journal={Matter},
  volume={1},
  number={5},
  pages={1370-1384},
  year={2019},
  publisher={Elsevier}
}

@article{court-et-al-voxel-vae,
  title={3-D inorganic crystal structure generation and property prediction via representation learning},
  author={Court, Callum J and Yildirim, Batuhan and Jain, Apoorv and Cole, Jacqueline M},
  journal={Journal of Chemical Information and Modeling},
  volume={60},
  number={10},
  pages={4518-4535},
  year={2020},
  publisher={ACS Publications}
}

@article{ZeoGAN,
  title={Inverse design of porous materials using artificial neural networks},
  author={Kim, Baekjun and Lee, Sangwon and Kim, Jihan},
  journal={Science advances},
  volume={6},
  number={1},
  pages={eaax9324},
  year={2020},
  publisher={American Association for the Advancement of Science}
}

@article{CCDCGAN,
  title={Constrained crystals deep convolutional generative adversarial network for the inverse design of crystal structures},
  author={Long, Teng and Fortunato, Nuno M and Opahle, Ingo and Zhang, Yixuan and Samathrakis, Ilias and Shen, Chen and Gutfleisch, Oliver and Zhang, Hongbin},
  journal={npj Computational Materials},
  volume={7},
  number={1},
  pages={66},
  year={2021},
  publisher={Nature Publishing Group UK London}
}

@article{FTCP,
  title={An invertible crystallographic representation for general inverse design of inorganic crystals with targeted properties},
  author={Ren, Zekun and Tian, Siyu Isaac Parker and Noh, Juhwan and Oviedo, Felipe and Xing, Guangzong and Li, Jiali and Liang, Qiaohao and Zhu, Ruiming and Aberle, Armin G and Sun, Shijing and others},
  journal={Matter},
  volume={5},
  number={1},
  pages={314-335},
  year={2022},
  publisher={Elsevier}
}

@article{CVAE,
  title={A scalable crystal representation for reverse engineering of novel inorganic materials using deep generative models},
  author={Bajpai, Rochan and Shukla, Atharva and Kumar, Janish and Tewari, Abhishek},
  journal={Computational Materials Science},
  volume={230},
  pages={112525},
  year={2023},
  publisher={Elsevier}
}

@article{wycryst,
	title={WyCryst: Wyckoff inorganic crystal generator framework},
	author={Zhu, Ruiming and Nong, Wei and Yamazaki, Shuya and Hippalgaonkar, Kedar},
	journal={Matter},
	volume={7},
	number={10},
	pages={3469--3488},
	year={2024},
	publisher={Elsevier}
}

@misc{cdvae,
      title={Crystal Diffusion Variational Autoencoder for Periodic Material Generation}, 
      author={Tian, Xie and Xiang, Fu and Octavian-Eugen, Ganea and Regina, Barzilay and Tommi, Jaakkola},
      year={2022},
      eprint={2110.06197},
      archivePrefix={arXiv},
      primaryClass={cs.LG},
      url={https://arxiv.org/abs/2110.06197}
}

@article{symat,
  title={Towards Symmetry-Aware Generation of Periodic Materials},
  author={Luo, Youzhi and Liu, Chengkai and Ji, Shuiwang},
  journal={Advances in Neural Information Processing Systems},
  volume={36},
  year={2024}
}

@article{dp-cdvae,
  title={Diffusion probabilistic models enhance variational autoencoder for crystal structure generative modeling},
  author={Pakornchote, Teerachote and Choomphon-Anomakhun, Natthaphon and Arrerut, Sorrjit and Atthapak, Chayanon and Khamkaeo, Sakarn and Chotibut, Thiparat and Bovornratanaraks, Thiti},
  journal={Scientific Reports},
  volume={14},
  number={1},
  pages={1275},
  year={2024},
  publisher={Nature Publishing Group UK London}
}

@article{con-cdvae,
  title={Con-CDVAE: A method for the conditional generation of crystal structures},
  author={Ye, Cai-Yuan and Weng, Hong-Ming and Wu, Quan-Sheng},
  journal={Computational Materials Today},
  pages={100003},
  year={2024},
  publisher={Elsevier}
}

@misc{cond-cdvae,
      title={Deep learning generative model for crystal structure prediction}, 
      author={Xiaoshan, Luo and Zhenyu, Wang and Pengyue, Gao and Jian, Lv and Yanchao, Wang and Changfeng, Chen and Yanming, Ma},
      year={2024},
      eprint={2403.10846},
      archivePrefix={arXiv},
      primaryClass={cond-mat.mtrl-sci},
      url={https://arxiv.org/abs/2403.10846}
}

@inproceedings{EMPNN,
  title={Equivariant message passing neural network for crystal material discovery},
  author={Klipfel, Astrid and Bouraoui, Zied and Peltre, Olivier and Fregier, Yael and Harrati, Najwa and Sayede, Adlane},
  booktitle={Proceedings of the AAAI Conference on Artificial Intelligence},
  volume={37},
  number={12},
  pages={14304--14311},
  year={2023}
}

@article{smld,
  title={Data-Driven Score-Based Models for Generating Stable Structures with Adaptive Crystal Cells},
  author={Sultanov, Arsen and Crivello, Jean-Claude and Rebafka, Tabea and Sokolovska, Nataliya},
  journal={Journal of Chemical Information and Modeling},
  volume={63},
  number={22},
  pages={6986-6997},
  year={2023},
  publisher={ACS Publications}
}

@article{diffcsp,
  title={Crystal structure prediction by joint equivariant diffusion},
  author={Jiao, Rui and Huang, Wenbing and Lin, Peijia and Han, Jiaqi and Chen, Pin and Lu, Yutong and Liu, Yang},
  journal={NeurIPS},
  volume={36},
  year={2024}
}

@article{mattergen,
	title={A generative model for inorganic materials design},
	author={Zeni, Claudio and Pinsler, Robert and Z{\"u}gner, Daniel and Fowler, Andrew and Horton, Matthew and Fu, Xiang and Wang, Zilong and Shysheya, Aliaksandra and Crabb{\'e}, Jonathan and Ueda, Shoko and others},
	journal={Nature},
	volume={639},
	number={8055},
	pages={624--632},
	year={2025},
	publisher={Nature Publishing Group UK London}
}

@inproceedings{gemsdiff,
  title={Vector Field Oriented Diffusion Model for Crystal Material Generation},
  author={Klipfel, Astrid and Fregier, Yael and Sayede, Adlane and Bouraoui, Zied},
  booktitle={Proceedings of the AAAI Conference on Artificial Intelligence},
  volume={38},
  number={20},
  pages={22193-22201},
  year={2024}
}

@article{diffcsp++,
  title={Space Group Constrained Crystal Generation}, 
  author={Rui, Jiao and Wenbing, Huang and Yu, Liu and Deli, Zhao and Yang, Liu},
  journal={arXiv:2402.03992},
  year={2024}
}

@article{symmcd,
	title={SymmCD: Symmetry-Preserving crystal generation with diffusion models},
	author={Levy, Daniel and Panigrahi, Siba Smarak and Kaba, S{\'e}kou-Oumar and Zhu, Qiang and Lee, Kin Long Kelvin and Galkin, Mikhail and Miret, Santiago and Ravanbakhsh, Siamak},
	journal={arXiv preprint arXiv:2502.03638},
	year={2025}
}

@article{wyckoffdiff,
	title={WyckoffDiff--A Generative Diffusion Model for Crystal Symmetry},
	author={Kelvinius, Filip Ekstr{\"o}m and Andersson, Oskar B and Parackal, Abhijith S and Qian, Dong and Armiento, Rickard and Lindsten, Fredrik},
	journal={arXiv preprint arXiv:2502.06485},
	year={2025}
}

@article{diffcrysgen,
	title={DiffCrysGen: A Score-Based Diffusion Model for Design of Diverse Inorganic Crystalline Materials},
	author={Mal, Sourav and Mishra, Subhankar and Sen, Prasenjit},
	journal={arXiv preprint arXiv:2505.07442},
	year={2025}
}

@article{Crystal-GFN,
  title = {Crystal-GFN: sampling crystals with desirable properties and constraints},
  author = {Hernandez-Garcia, Alex and Duval, Alexandre and Volokhova, Alexandra and Bengio, Yoshua and Sharma, Divya and Carrier, Pierre Luc and Koziarski, Michal and Schmidt, Victor},
  journal={Proceedings of NeurIPS 2023},
  year={2023}
}

@inproceedings{CHGFlowNets,
  title={Hierarchical GFlownet for Crystal Structure Generation},
  author={Nguyen, Tri Minh and Tawfik, Sherif Abdulkader and Tran, Truyen and Gupta, Sunil and Rana, Santu and Venkatesh, Svetha},
  booktitle={AI for Accelerated Materials Design-NeurIPS 2023 Workshop},
  year={2023}
}

@article{LLaMA-2-fine-tuned,
      title={Fine-Tuned Language Models Generate Stable Inorganic Materials as Text}, 
      author={Nate, Gruver and Anuroop, Sriram and Andrea, Madotto and Andrew, Gordon Wilson and C. Lawrence, Zitnick and Zachary, Ulissi},
      journal={arXiv:2402.04379},
      year={2024}
}

@article{SLICES,
  title={An invertible, invariant crystal representation for inverse design of solid-state materials using generative deep learning},
  author={Xiao, Hang and Li, Rong and Shi, Xiaoyang and Chen, Yan and Zhu, Liangliang and Chen, Xi and Wang, Lei},
  journal={Nature Communications},
  volume={14},
  number={1},
  pages={7027},
  year={2023},
  publisher={Nature Publishing Group UK London}
}

@article{CrystalFormer,
	title={Space Group Informed Transformer for Crystalline Materials Generation}, 
	author={Zhendong, Cao and Xiaoshan, Luo and Jian, Lv and Lei, Wang},
	journal={arXiv:2403.15734},
	year={2024}
}

@article{CDI_CDS,
	title={Generative Inverse Design of Crystal Structures via Diffusion Models with Transformers},
	author={Takahara, Izumi and Shibata, Kiyou and Mizoguchi, Teruyasu},
	journal={arXiv preprint arXiv:2406.09263},
	year={2024}
}

@article{Uni-MDM,
	title={A Generation Framework with Strict Constraints for Crystal Materials Design},
	author={Huang, Chao and Chen, Jiahui and Chen, Chen and Chen, Chunyan and Su, Renjie and Du, Shiyu and Liang, Hongrui and Lin, Daojing and others},
	journal={arXiv preprint arXiv:2411.08464},
	year={2024}
}

@article{wyformer,
	title={Wyckoff transformer: Generation of symmetric crystals},
	author={Kazeev, Nikita and Nong, Wei and Romanov, Ignat and Zhu, Ruiming and Ustyuzhanin, Andrey and Yamazaki, Shuya and Hippalgaonkar, Kedar},
	journal={arXiv preprint arXiv:2503.02407},
	year={2025}
}

@article{ADiT,
	title={All-atom diffusion transformers: Unified generative modelling of molecules and materials},
	author={Joshi, Chaitanya K and Fu, Xiang and Liao, Yi-Lun and Gharakhanyan, Vahe and Miller, Benjamin Kurt and Sriram, Anuroop and Ulissi, Zachary W},
	journal={arXiv preprint arXiv:2503.03965},
	year={2025}
}

@article{CrystalDiT,
  title={CrystalDiT: A Diffusion Transformer for Crystal Generation},
  author={Yi, Xiaohan and Xu, Guikun and Xiao, Xi and Zhang, Zhong and Liu, Liu and Bian, Yatao and Zhao, Peilin},
  journal={arXiv preprint arXiv:2508.16614},
  year={2025}
}

@article{Yang2019ComplexTransformer,
  title        = {Complex Transformer: A Framework for Modeling Complex-Valued Sequence},
  author       = {Muqiao Yang and Martin Q. Ma and Dongyu Li and Yao-Hung Hubert Tsai and Ruslan Salakhutdinov},
  journal      = {ArXiv},
  year         = {2019},
  volume       = {abs/1910.10202},
  url          = {https://arxiv.org/abs/1910.10202}
}

@article{Eilers2023BuildingBlocksCV,
  title        = {Building Blocks for a Complex-Valued Transformer Architecture},
  author       = {Florian Eilers and Xiaoyi Jiang},
  journal      = {ArXiv},
  year         = {2023},
  volume       = {abs/2306.09827},
  url          = {https://arxiv.org/abs/2306.09827}
}

@article{vaswani2017attention,
  title={Attention is all you need},
  author={Vaswani, Ashish and Shazeer, Noam and Parmar, Niki and Uszkoreit, Jakob and Jones, Llion and Gomez, Aidan N and Kaiser, {\L}ukasz and Polosukhin, Illia},
  journal={Advances in neural information processing systems},
  volume={30},
  year={2017}
}

@article{Su2021RoPE,
  title   = {RoFormer: Enhanced Transformer with Rotary Position Embedding},
  author  = {Su, Jianlin and Lu, Yu and Pan, Shengfeng and Wen, Bo and Liu, Yunfeng},
  journal = {arXiv preprint arXiv:2104.09864},
  year    = {2021},
  url     = {https://arxiv.org/abs/2104.09864}
}

@inproceedings{ronneberger2015u,
  title={U-net: Convolutional networks for biomedical image segmentation},
  author={Ronneberger, Olaf and Fischer, Philipp and Brox, Thomas},
  booktitle={International Conference on Medical image computing and computer-assisted intervention},
  pages={234--241},
  year={2015},
  organization={Springer}
}

@inproceedings{jaegle2021perceiver,
  title={Perceiver: General perception with iterative attention},
  author={Jaegle, Andrew and Gimeno, Felix and Brock, Andy and Vinyals, Oriol and Zisserman, Andrew and Carreira, Joao},
  booktitle={International conference on machine learning},
  pages={4651--4664},
  year={2021},
  organization={PMLR}
}

@misc{lematerial_2025,
  author       = {{LeMaterial Consortium}},
  title        = {LeMaterial: Large-Scale Materials Datasets for Machine Learning},
  year         = {2025},
  howpublished = {\url{https://huggingface.co/datasets/LeMaterial}},
  note         = {\url{https://huggingface.co/datasets/LeMaterial/LeMat-Bulk}},
}

@misc{lemat-bulkunique,
  author={Martin Siron and Inel Djafar and Lucile Ritchie and Etienne Du-Fayet and Amandine Rossello and Ali Ramlaoui and Leandro von Werra and Thomas Wolf and Alexandre Duval},
  title={ LeMat-BulkUnique Dataset },
  year={2024},
  url={ https://huggingface.co/datasets/LeMaterial/LeMat-BulkUnique },
  publisher={ Hugging Face }
}

@article{materials_project,
    author = {Jain, Anubhav and Ong, Shyue Ping and Hautier, Geoffroy and Chen, Wei and Richards, William Davidson and Dacek, Stephen and Cholia, Shreyas and Gunter, Dan and Skinner, David and Ceder, Gerbrand and Persson, Kristin A.},
    title = {Commentary: The Materials Project: A materials genome approach to accelerating materials innovation},
    journal = {APL Materials},
    volume = {1},
    number = {1},
    pages = {011002},
    year = {2013},
    month = {07},
    issn = {2166-532X},
    doi = {10.1063/1.4812323},
    url = {https://doi.org/10.1063/1.4812323},
    eprint = {https://pubs.aip.org/aip/apm/article-pdf/doi/10.1063/1.4812323/13163869/011002_1_online.pdf},
}

@article{alexandria_pbe,
author = {Jonathan Schmidt  and Love Pettersson  and Claudio Verdozzi  and Silvana Botti  and Miguel A. L. Marques },
title = {Crystal graph attention networks for the prediction of stable materials},
journal = {Science Advances},
volume = {7},
number = {49},
pages = {eabi7948},
year = {2021},
doi = {10.1126/sciadv.abi7948},
URL = {https://www.science.org/doi/abs/10.1126/sciadv.abi7948},
eprint = {https://www.science.org/doi/pdf/10.1126/sciadv.abi7948},
}

@article{oqmd,
  title={Materials design and discovery with high-throughput density functional theory: the open quantum materials database (OQMD)},
  author={Saal, James E and Kirklin, Scott and Aykol, Muratahan and Meredig, Bryce and Wolverton, Christopher},
  journal={Jom},
  volume={65},
  number={11},
  pages={1501--1509},
  year={2013},
  publisher={Springer}
}

@bool{international_tables_crystallography,
  title     = "International Tables for Crystallography, {Space-Group} Symmetry",
  author    = "Aroyo, M I",
  publisher = "Wiley",
  year      =  2017
}

@article{Mroz2022,
author={Mroz, Austin M.
and Posligua, Victor
and Tarzia, Andrew
and Wolpert, Emma H.
and Jelfs, Kim E.},
title={Into the Unknown: How Computation Can Help Explore Uncharted Material Space},
journal={Journal of the American Chemical Society},
year={2022},
month={Oct},
day={19},
publisher={American Chemical Society},
volume={144},
number={41},
pages={18730-18743},
issn={0002-7863},
doi={10.1021/jacs.2c06833},
url={https://doi.org/10.1021/jacs.2c06833}
}
\bibliographystyle{icml2026}

\appendix

\section*{Appendix Overview}
This appendix provides additional technical details, derivations, and extended
experimental results that support the main text. The material is organized to
mirror the structure of the paper and to facilitate reproducibility without
interrupting the narrative flow of the main body.

\section{Dataset and Preprocessing Details}
\label{app:dataset}

\subsection{Source datasets and harmonization}
\label{app:dataset_sources}

\noindent\textbf{LeMaterial}
\label{app:LeMaterial}
The \emph{LeMaterial initiative}~\citep{lematerial_2025} is a recent effort aimed at
curating large-scale, high-quality open datasets and developing tooling to
support machine-learning research in materials science. Its first major release,
\emph{LeMat-Bulk}, aggregates crystal structures computed using density functional
theory (DFT) from three widely used public databases: the Materials Project~\citep{materials_project},
Alexandria~\citep{alexandria_pbe}, and the Open Quantum Materials Database
(OQMD)~\citep{oqmd}.

In this work, we use the PBE subset of the \emph{LeMat-BulkUnique} dataset~\citep{lemat-bulkunique},
which contains $5{,}005{,}017$ unique crystal structures as determined by a
fingerprinting procedure developed by the LeMaterial team. All structures in
this subset were computed using the Perdew--Burke--Ernzerhof (PBE)
exchange--correlation functional~\citep{pbe_functional}, which remains the most
commonly used functional for large public crystal databases due to its favorable
trade-off between accuracy and computational efficiency. We refer to this dataset
hereafter as \emph{LeMat-BulkUnique-PBE}.

\noindent\textbf{Pre-filters and splits}
\label{app:prefilters-and-splits}
We apply a sequence of standard pre-filters to \emph{LeMat-BulkUnique-PBE} in order
to obtain a clean and harmonized corpus suitable for large-scale pretraining.
Specifically, we remove structures containing noble gases or \emph{f}-block
elements, structures with interatomic distances smaller than
\qty{0.5}{\angstrom}, and duplicate structures identified using the
\texttt{StructureMatcher} from \textsc{pymatgen}, with fractional length tolerance
$\texttt{ltol}=0.2$, site tolerance $\texttt{stol}=0.3$, and angular tolerance
$\texttt{angle-tol}=\qty{5}{\degree}$. These tolerance values follow commonly used
defaults in prior work and are intended to remove only near-identical duplicates.

After filtering, the resulting dataset contains $2{,}838{,}937$ structures and is
publicly released in CIF format on HuggingFace\footnote{\url{https://huggingface.co/datasets/materials-toolkits/lematerial}}.
We refer to this corpus as the \emph{LeMat-pretrain} dataset.

From \emph{LeMat-pretrain}, we further extract a subset of at least metastable
structures, defined as having an energy above the convex hull less than or equal
to \qty{0.1}{\evperatom}. Energies above hull are obtained from the
\emph{LeMat-Bulk-DFT-Hull-All} dataset\footnote{\url{https://huggingface.co/datasets/LeMaterial/LeMat-Bulk-DFT-Hull-All}},
using the provided \texttt{dft-hull} values. This metastable subset contains
$506{,}853$ structures and is referred to as \emph{LeMat-metastable}.

\noindent\textbf{Train--test split}
From the metastable subset, we reserve $2{,}048$ structures for testing.
To ensure coverage across structural complexity, the test set is stratified by
unit-cell size: $512$ structures with $\leq 16$ atoms, $512$ with $17$--$32$ atoms,
$512$ with $33$--$48$ atoms, and $512$ with $49$--$64$ atoms.
The remaining $504{,}805$ metastable structures are used for training.

\subsection{Fractional-coordinate snapping and rational grids}
\label{app:snapping}
Fractional atomic coordinates are snapped to a rational grid defined by an
integer denominator, such as $24$ or $48$. This discretization facilitates stable
Fourier representations while preserving crystallographic symmetries.
In fractional coordinates, the admissible translational components that can
combine with integer rotation matrices to form lattice isomorphisms are drawn
from the finite set $\left\{\,0,\;\tfrac{1}{6},\;\tfrac{1}{4},\;\tfrac{1}{3},\;\tfrac{1}{2}\right\}$ \cite{international_tables_crystallography}.
By choosing a snapping denominator that is a multiple of $12$, all such symmetry-%
compatible translations arrive at the same rounding error on the snapping grid.
As a result, symmetry-equivalent atoms are rounded coherently, ensuring that
snapping does not break crystallographic equivalence relations.

In this work, $48$ is the largest snapping denominator used, corresponding to a
maximum roundoff error of $1/96$, i.e.\ approximately $1\%$ in fractional
coordinates. This bound provides a lightweight justification that the induced
quantization error remains small relative to typical structural variations while
enabling exact recovery on the chosen grid.

\subsection{Recoverability criteria and rejection conditions}
\label{app:recoverability}
The fractional-coordinate recovery procedure is described in detail in
Appendix~\ref{app:reconstruction}. In brief, recovery is attempted in a sequence of
up to three stages of increasing numerical precision and computational cost. If
recovery fails at a given stage, the procedure advances to the next stage.

Because fractional coordinates are snapped to a rational grid, we require exact
recovery of all fractional coordinates for all atomic species present. Crystals
for which exact recovery fails after the final stage are rejected. This criterion
is applied prior to training and ensures that Fourier coefficients can be mapped
unambiguously back to fractional coordinates, allowing reconstruction accuracy to
be assessed without storing explicit coordinates alongside the Fourier
representation.

\section{Representation and Symmetry Structure}
\label{app:fourier}

\subsection{Rotation-invariant lattice parameterization}
\label{app:lattice_param}

Material properties derived from a crystalline structure are invariant under
global rotations of the crystallographic axes. Accordingly, we adopt a lattice
representation that factors out rotational degrees of freedom while remaining
well conditioned across a wide range of length scales.

Let $\mLatt \in \sReal^{3\times3}$ denote the lattice matrix. We obtain a
rotation-invariant encoding via the polar decomposition,
\[
\mLatt = \mR \exp(\mSymm),
\]
where $\mR \in \mathrm{SO}(3)$ is a rotation and $\exp(\mSymm) \in
\mathrm{Sym}^+(3)$ is symmetric positive definite. Equivalently, writing the
singular value decomposition $\mLatt = \mU \mSigma \mV^\top$ with
$\mSigma = \diag(\vSigma)$, the polar factors are given by
\[
\mR = \mU \mV^\top,
\qquad
\mSymm = \mV \diag(\log \vSigma)\, \mV^\top.
\]

The matrix $\mSymm$ provides an additive, rotation-invariant parameterization of
lattice geometry via the matrix logarithm of the symmetric polar factor, a
construction that is standard in lattice analysis and geometry processing
\cite{}. Isotropic rescaling corresponds to shifts proportional to the identity,
while anisotropic deformations are encoded by traceless components. This
parameterization is therefore well suited for learning and optimization across
diverse lattice scales.

\subsection{Choice of truncation geometry}
\label{app:truncation_geometry}

The truncated Fourier representation requires selecting a finite subset of
reciprocal-space wave vectors $\sWave$. While spherical truncation
($\|\vWave\|_2 \le \jMax$) is analytically natural for isolated atoms or radially
symmetric densities, we find it poorly matched to crystalline configurations,
which often exhibit separable structure along lattice-aligned directions.

Across a large and diverse corpus of crystalline materials, cubic truncation
($\|\vWave\|_\infty \le \jMax$) consistently yields higher reconstruction success
rates than spherical or hybrid truncation schemes, even in cases where the latter
retain a larger number of Fourier modes. We therefore adopt cubic truncation
throughout this work as a practical and empirically robust choice.

\noindent\textbf{Reconstruction limits under cubic truncation}
Using cubic truncation, reconstruction success varies smoothly with atomic count
and unit-cell occupancy. For $7^3$ retained modes ($\jMax = 3$), we observe at
least one crystal containing $52$ atoms of a single species that can be fully
reconstructed from its truncated Fourier representation. Increasing the
resolution to $9^3$ modes ($\jMax = 4$) similarly permits reconstruction of at
least one crystal containing $108$ atoms of a single species. These observations
illustrate the scaling behavior of recoverability under cubic truncation and
motivate the use of higher Fourier resolution when targeting structurally complex
or high-occupancy unit cells.

\subsection{Fourier-space symmetries and algebraic constraints}
\label{app:symmetry_algebra}

Crystallographic symmetries act on atomic configurations as isomorphisms generated
by compositions of rotations and translations that preserve the lattice \cite{EMPNN,mattergen}. When expressed in fractional coordinates, admissible rotational
symmetries are represented by integer-valued matrices
$\mM \in \sInt^{3\times3}$ with $\det \mM = \pm 1$, corresponding to lattice
automorphisms. Associated symmetry translations
$\vDelta \in \sReal^3$ are restricted to rational fractional shifts consistent
with the space group of the crystal.

Under such a symmetry operation, a fractional atomic coordinate
$\vFrac \in \sUnit_{\nSpec}$ transforms as
\[
\vFrac' = \mM \vFrac + \vDelta \pmod{1}.
\]
Substituting this transformation into the definition of the truncated Fourier
coefficients yields an induced action on reciprocal-space components. In
particular, the Fourier coefficients for species $\nSpec$ satisfy
\[
\vFour_{\vWave}^{(\nSpec)} =
\exp\!\left(-2\pi i\, \vWave^\top \vDelta\right)\,
\vFour_{\mM^\top \vWave}^{(\nSpec)},
\]
where the rotation permutes wave vectors via
$\vWave \mapsto \mM^\top \vWave$ and the translation contributes a phase factor.

These algebraic constraints encode crystallographic symmetry directly in
reciprocal space, revealing structured redundancies among Fourier coefficients.
In the proposed representation, such redundancies can be learned, compressed, and
respected implicitly in the latent space, without requiring explicit symmetry
enforcement during generation.

\section{Tokenization and Embedding Details}
\label{app:tokenization_overview}

This section specifies the tokenization and embedding scheme summarized in the
main text, describing how lattice geometry, species identity, and Fourier
coefficients are embedded into a unified complex-valued token sequence.

\subsection{Token construction and embedding}
\label{app:tokenization}

The transformer input consists of a small set of global descriptors together
with a structured collection of complex Fourier coefficients representing the
atomic density. All components are embedded into a common complex vector space
$\mathbb{C}^{\dModel}$ prior to transformer processing.

Global descriptors comprise the rotation-invariant lattice parameters described
in Appendix~\ref{app:lattice_param} and the identities of the atomic species
present in the structure. Local structural information is represented by one
token per retained reciprocal-space wave vector.

\subsection{Species-slot assignment and randomization}
\label{app:species_slots}

The architecture supports up to six distinct atomic species per structure. This
limit is dataset-driven and synthesis-practical, excluding fewer than $0.01\%$
of structures while substantially simplifying token routing and improving
compression efficiency.

Species identities are represented using learned embeddings without imposing
explicit chemical priors. When a structure is drawn from the training set, the
assignment of species to the six available slots is randomized: a nonzero column
is selected uniformly at random, and remaining species are placed consecutively
modulo six. This permutation is applied coherently across all Fourier rows so
that each species occupies a single column throughout the representation. The
procedure prevents systematic undertraining of specific slots while preserving
species consistency across tokens.

\subsection{Global structure token construction}
\label{app:global_token}

For each structure, the six lattice parameters and the embeddings corresponding
to the species present (ordered by atomic number, consistent with the Fourier
representation) are concatenated into a single real-valued vector. This vector
is mapped through a learned linear projection from
$\mathbb{R}^{6(1 + d_{\text{enc}})}$ to $2\dModel$ real channels, which are
interpreted as a complex vector in $\mathbb{C}^{\dModel}$. The result is a single
global token encoding lattice geometry and species composition.

\subsection{Fourier token projection and parameterization}
\label{app:fourier_tokens}

The truncated Fourier representation is organized as a matrix of complex
coefficients, with rows indexed by reciprocal-space wave vectors and columns by
species slots. Species absent from a structure have identically zero
coefficients in the corresponding columns.

The Fourier rows, each represented as a vector in $\mathbb{C}^6$,
are also mapped to complex tokens, one per wave vector,
by a complex-valued linear transformation.
Using a single shared projection enforces consistent parameterization across reciprocal space.
Because the input is already complex-valued, this projection halves the number of
learnable parameters relative to an equivalent real-valued projection with
doubled channel width, while preserving the total number of real-valued floating-
point operations.

\noindent\textbf{Auxiliary tokens and capacity control}
\label{app:aux_tokens}

A small block of auxiliary complex tokens with learned initial values is
prepended to the token sequence. These tokens provide the depth-aligned
scaffold that becomes the auxiliary ladder encoding.
The number of auxiliary tokens is treated as a hyperparameter and is chosen in
proportion to Fourier resolution (e.g., $5$ auxiliary tokens for $7^3$ retained
modes and $9$ for $9^3$). Auxiliary tokens are initialized from trainable
parameters and introduced once at the input.

\section{Complex Transformer Blocks}
\label{app:complex_transformer}

Each layer of the encoder and decoder consists of a standard transformer block
adapted to operate natively on complex-valued token representations. The overall
structure follows a pre-norm normalization–attention–MLP pipeline with residual
connections applied at each sublayer, using complex-valued linear maps and
geometry-aware positional encoding.

\subsection{Complex RMS normalization}
\label{app:rmsnorm}

Let $X \in \mathbb{C}^{S \times \dModel}$ denote the input token sequence. We apply
a complex-valued RMS normalization based on the mean squared modulus of the
activations,
\[
\hat X = \frac{X}{\sqrt{\mathbb{E}[|X|^2] + \varepsilon}},
\]
where the expectation is taken over the feature dimension. This normalization is
independent of the complex phase of the input and rescales each token uniformly.
In practice, $X$ is represented as a real tensor of dimension $2\dModel$, and the
mean of squared real components is computed, differing from real RMS normalization
by only a constant factor.

After normalization, a learned complex affine transformation is applied,
\[
\mathrm{RMSNorm}(X) = \hat X \odot w + b,
\]
with complex-valued scale $w \in \mathbb{C}^{\dModel}$ and bias
$b \in \mathbb{C}^{\dModel}$. Magnitude-based normalization in complex space is
consistent with prior complex transformer designs
\citep{Yang2019ComplexTransformer,Eilers2023BuildingBlocksCV}. We include a complex
bias term, which empirically improves VAE optimization (see ablation study).

\subsection{Complex multi-head attention and real-valued kernels}
\label{app:complex_attention}

Queries, keys, and values are produced via complex linear projections of the
normalized activations,
\[
Q = \hat X W_Q, \quad K = \hat X W_K, \quad V = \hat X W_V,
\]
with $W_Q,W_K,W_V \in \mathbb{C}^{\dModel \times \dModel}$. These tensors are
reshaped into multiple heads in the standard manner.

To ensure that attention weights form convex combinations of values, attention
logits are computed from a real-valued similarity measure. Specifically, the
attention score between a query $q$ and key $k$ is given by
\[
\mathrm{score}(q,k) = \Re\!\left(q^\top k^\ast\right).
\]
In implementation, real and imaginary components are packed as additional feature
dimensions and standard scaled dot-product attention (SDPA) kernels are applied
to the resulting real-valued tensors. This yields real, properly normalized
attention weights while preserving compatibility with optimized attention
implementations.

\subsection{Complex Fourier rotational positional encoding (RoPE3D)}
\label{app:rope3d}

Positional information is incorporated using a complex-valued rotational
positional encoding (RoPE), extending the formulation of
\citet{Su2021RoPE} to three-dimensional reciprocal space. Each attention head of
dimension $d_{\text{head}}$ is constructed such that $d_{\text{head}}$ is divisible
by three and is partitioned into equal subspaces corresponding to the $x$, $y$,
and $z$ components of a reciprocal-space wave vector.

For Fourier tokens, base rotation angles are deterministically derived from the
associated wave vectors using a fixed set of geometrically scaled frequencies.
Auxiliary and global tokens, which do not correspond to specific reciprocal-space
locations, receive zero base angles. In all cases, trainable per-head and
per-channel phase offsets are added, allowing the model to adapt the positional
encoding during training.

The resulting rotations are applied elementwise to queries and keys via
multiplication by $\exp(i\theta)$, where $\theta$ denotes the sum of base and
learned angles. Because representations are complex-valued, positional encoding
reduces to a simple complex phase rotation.

\subsection{Complex gated MLP}
\label{app:complex_mlp}

Following the attention sublayer, a complex-valued gated MLP is applied. We
consider two gating formulations that differ in how nonlinear modulation is
applied to complex activations.

\noindent\textbf{Separate real--imaginary gating (baseline)}
In the baseline architecture, activations are complex RMS-normalized and
projected into two complex-valued streams, $U$ and $G$, using complex linear
maps. Gating is applied by interpreting these tensors in their real-valued
representation: a SiLU nonlinearity is applied independently to the real and
imaginary components of $G$, and the result is multiplied elementwise with the
corresponding components of $U$. The gated output is then reassembled into a
complex tensor and passed through a final complex linear projection. Optional
dropout is applied to the update stream prior to gating.

This formulation provides flexible component-wise modulation but does not
explicitly preserve phase structure under nonlinear gating.

\noindent\textbf{Complex modulus gating (phase-preserving variant)}
In the modulus-gated variant, activations are complex RMS-normalized and
projected into a single complex update stream $U$. A real-valued gate $G$ is
computed from the concatenated real and imaginary components of the normalized
activations using a real linear projection followed by a SiLU nonlinearity.
Dropout is applied independently to the real and imaginary parts of $U$ prior to
gating. The gated activation is then formed by modulating the complex update
stream with the real-valued gate, $U \odot G$, and projected back to $\dModel$.

Because the gate acts multiplicatively on the activation magnitude, this
formulation modulates amplitude while preserving relative phase information
across complex channels. As shown in the ablation study
(Section~\ref{sec:ablation}), this variant reduces computational overhead and
improves optimization efficiency relative to separate real--imaginary gating.

\subsection{Residual composition}
\label{app:residuals}

Attention and MLP sublayers are combined using standard pre-norm residual
connections. Given an input $X$, the block computes
\begin{align*}
Y &= X + \mathrm{Attention}(\mathrm{RMSNorm}(X)),\\
Z &= Y + \mathrm{MLP}(\mathrm{RMSNorm}(Y)),
\end{align*}
with all operations performed in complex space. This yields a direct complex-valued
analogue of the standard transformer block \citep{Vaswani2017Attention}, while
preserving compatibility with efficient real-valued kernels where appropriate.

\section{Auxiliary Ladder and Channel Suppression}
\label{app:aux_ladder}

This section describes the auxiliary ladder compression mechanism, including
channel-wise masking, signal-to-noise regularization, and the nonzero-element
(nnz) scheduling strategy used to suppress and prune latent channels during
training.

\subsection{Channel-wise masking and selector dynamics}
\label{app:selector}

The auxiliary ladder is formed by concatenating auxiliary token states extracted
at each encoder depth, yielding a depth-aligned latent representation
$\vEnc \in \mathbb{C}^{(L n_{\text{aux}})\times \dModel}$. To encourage compression
and enable adaptive channel pruning, we apply a learned real-valued mask of the
same shape as the auxiliary ladder, with one scalar mask parameter per complex
channel. Each mask element jointly scales the real and imaginary components of its
associated channel, thereby modulating the channel magnitude.

Mask parameters are trained under a signal-to-noise regularization scheme that
penalizes the $\ell_1$ norm of the ratio between channel magnitudes and their
corresponding learned noise scales. This formulation yields constant-magnitude
gradients with respect to channel amplitude per unit noise, even as
activations approach zero, preventing early gradient collapse and enabling
principled suppression of non-informative channels. Adaptive, channel-specific
noise scales were found to substantially improve optimization stability compared
to a shared noise level.

Channels whose masked magnitudes are driven sufficiently close to zero are
eligible for permanent removal through the nnz scheduling mechanism described
below, producing a binary selector that deactivates pruned channels.

\subsection{Latent signal-to-noise regularization}
\label{app:snr_latent}

Latent regularization is implemented directly on the auxiliary ladder via an
explicit signal-to-noise ratio (SNR) penalty, rather than through a Gaussian KL
divergence. Given the auxiliary ladder mean $\vEnc$ and learned per-channel
log-scale $\vSigma$, the penalty is defined as
\begin{equation}
\mathcal{L}_{\mu} =
\frac{1}{L n_{\text{aux}}\, \dModel}
\sum \frac{|\vEnc|}{\exp(\vSigma)},
\end{equation}
where $|\vEnc| = \sqrt{\Re(\vEnc)^2 + \Im(\vEnc)^2}$ denotes the complex modulus and
the sum runs over all elements of the auxiliary ladder.

This SNR-based penalty maintains nonvanishing gradient pressure in the
low-amplitude regime, in contrast to Gaussian KL regularization whose gradient
vanishes near zero. As a result, it promotes learned channel suppression and
provides a continuous signal that complements the discrete pruning induced by
masking and nnz scheduling.

\subsection{Nonzero-element scheduling and pruning}
\label{app:nnz_schedule}

In addition to continuous suppression, we employ an explicit nonzero-element
(nnz) scheduling mechanism to permanently prune latent channels. Here, a
\emph{channel} refers to a single complex element of a single auxiliary token;
channels are not shared across tokens or depths.

Training begins with all auxiliary ladder channels active. Over a prescribed
number of steps, the target nnz count is annealed from the full channel set to a
final value using a cosine schedule. At each pruning step, channels are ranked by
the magnitude of their learned mask values, and the lowest-ranked channels are
irreversibly dropped to meet the current nnz target. Dropped channels are
permanently deactivated and do not re-enter training.

The final nnz target and the duration of the cosine schedule are treated as
hyperparameters and are varied in the extended sensitivity experiments reported
in Appendix~\ref{app:compression_sensitivity}. This combination of continuous
suppression and scheduled pruning enables efficient capacity allocation while
maintaining stable optimization.

\section{Training Objectives and Optimization}
\label{app:training}

\subsection{Scale-Invariant Reconstruction Loss}
\label{app:recon_aggregation}

Direct summation of reconstruction losses yields gradients proportional to absolute
error magnitude, causing optimization pressure to decay rapidly as reconstruction
improves. To maintain stable and balanced gradients throughout training, lattice
and Fourier reconstruction losses are combined using an $\ell_2$ aggregation,
\begin{equation}
\lRec = \sqrt{\lLat + \lFour}.
\end{equation}
This corresponds to an $\ell_2$ norm over stacked reconstruction residuals and
yields gradients that are normalized in direction and weakly dependent on overall
error scale. As a result, the relative weighting between lattice and Fourier
reconstruction remains stable across training, without requiring manual
rebalancing.

\subsection{Latent regularization and relation to KL divergence}
\label{app:latent_regularization}

Latent regularization is implemented using the signal-to-noise–based penalty
defined in Appendix~\ref{app:snr_latent}, rather than a Gaussian KL divergence.
This choice is motivated by optimization behavior rather than a departure from
the variational framework.

Under a diagonal complex Gaussian posterior with mean $\vEnc$ and variance
$\exp(2\vSigma)$, the KL divergence to a zero-mean prior with matched variance is
proportional to the squared normalized mean,
\[
\mathrm{KL} \;\propto\; \sum |\vEnc|^2/\exp(2\vSigma).
\]
The SNR penalty replaces this quadratic dependence with a monotonic function of
the same quantity,
\[
\sum |\vEnc|/\exp(\vSigma),
\]
which preserves the ordering induced by KL divergence while maintaining
approximately constant gradient pressure as channel amplitudes approach zero.

From an information-theoretic perspective, this can be viewed as optimizing a
monotonic surrogate of the KL divergence that emphasizes suppressibility and
channel selection over strict moment matching. This modification avoids the
vanishing gradients of Gaussian KL regularization in the low-amplitude regime,
enabling learned channel suppression and effective interaction with the masking
and nnz pruning mechanisms described in Appendix~\ref{app:aux_ladder}.

\subsection{Optimization schedules and weight decay}
\label{app:optimization}

All models are trained using Adam optimization. We use an initial learning rate
of $2\times10^{-4}$, with linear warmup from $10^{-7}$ over the first $1{,}000$
steps, followed by cosine decay to $2\times10^{-5}$ over a single training epoch
(approximately $3.3\times10^{5}$ steps). Light weight decay ($10^{-9}$) is applied
during warmup only and disabled thereafter. No gradient clipping or mixed-precision
training is used.

\section{Multistage Reconstruction of Atomic Coordinates}
\label{app:reconstruction}

After the final decoder block, linear reconstruction heads map the decoded global
structure token to lattice parameters and species logits (one slot per species),
and map the decoded Fourier tokens to species-resolved Fourier coefficients for
each retained reciprocal-space wave vector. Fractional atomic coordinates are
then recovered from the Fourier representation using a multistage procedure
described below.

Reconstruction is performed independently for each atomic species. The number of
atoms $n_{\nSpec}$ of a given species is determined \emph{a priori} from the
zero-frequency Fourier coefficient $\vFour_{\vWave=\bm{0}}$. Recovery proceeds in
three stages of increasing precision and computational cost. If reconstruction
fails after the final stage, the structure is rejected.

\subsection{Stage~1: Direct density-peak recovery}
\label{app:recon_stage1}

We evaluate the reconstructed atomic density on a discrete fractional-coordinate
grid matching the rational grid used during preprocessing (e.g., $48^3$ grid
points for denominator~48),
\[
\rho(\vFrac)
=
\sum_{\vWave \in \sWave}
\vFour_{\vWave}\,
\exp\!\left(2\pi i\, \vWave^\top \vFrac\right),
\]
and select the $n_{\nSpec}$ grid points with maximal density. When fractional
coordinates have been pre-rounded to this grid, these maxima typically coincide
with the original rounded coordinates. The majority of structures are fully
recovered at this stage.

\subsection{Stage~2: Iterative atom removal}
\label{app:recon_stage2}

If direct density-peak recovery fails to identify all atoms, coordinates are
recovered sequentially using a greedy procedure. After identifying a density
maximum at $\vFrac_a$, its exact Fourier contribution,
\[
\exp(-2\pi i\, \vWave^\top \vFrac_a),
\]
is subtracted from the Fourier coefficients, assuming unit occupancy. The updated
density is then recomputed and the next maximum located. This procedure is
repeated until all $n_{\nSpec}$ atoms are recovered or no further maxima can be
identified. In practice, this stage resolves nearly all remaining cases.

\subsection{Stage~3: Newton refinement and least-squares correction}
\label{app:recon_stage3}

In rare cases, nearby atoms may shift density maxima or lead to coincident grid
locations. To resolve such ambiguities, we introduce a small random perturbation
to all recovered coordinates and jointly refine the full set of atomic positions
for a given species by minimizing the residual between the target Fourier
coefficients and those induced by the reconstructed coordinates.

Let $\sUnit = \{\vFrac_a\}$ denote the current set of fractional coordinates for a
species, and let $\mathcal{F}(\sUnit)$ denote the corresponding truncated Fourier
coefficients,
\[
\mathcal{F}_{\vWave}(\sUnit)
=
\sum_{\vFrac_a \in \sUnit}
\exp(-2\pi i\, \vWave^\top \vFrac_a).
\]
The partial derivatives with respect to each coordinate are
\[
\frac{\partial \mathcal{F}_{\vWave}}{\partial \vFrac_a}
=
-2\pi i\, \vWave\,
\exp(-2\pi i\, \vWave^\top \vFrac_a),
\]
which form a Jacobian matrix $\mD$ with columns corresponding to the flattened
fractional coordinates in $\sUnit$.

To maintain real-valued coordinates, we construct an augmented real system,
\[
\begin{bmatrix}
\Re(\mathcal{F}(\sUnit + \Delta))\\
\Im(\mathcal{F}(\sUnit + \Delta))
\end{bmatrix}
\approx
\begin{bmatrix}
\Re(\mathcal{F}(\sUnit))\\
\Im(\mathcal{F}(\sUnit))
\end{bmatrix}
+
\begin{bmatrix}
\Re(\mD)\\
\Im(\mD)
\end{bmatrix}
\Delta,
\]
and solve the resulting least-squares problem using a standard SVD-based
pseudoinverse,
\[
\Delta
=
\mD^\dagger
\bigl[
\mathcal{F}(\sUnit) - \vFour
\bigr].
\]
Coordinates are updated as $\sUnit \gets \sUnit - \Delta$ and the procedure is
iterated up to a maximum of $10$ iterations. If convergence is not achieved
within this limit, reconstruction is deemed to have failed and the structure is
rejected.

Crystallographic symmetry is handled implicitly through the Fourier
representation: symmetry-related atoms induce identical constraints in reciprocal
space, and no explicit symmetry enforcement is required during reconstruction.

\section{Reverse Diffusion Formulation}
\label{app:diffusion}

This section describes the reverse diffusion formulation used to generate latent
auxiliary ladders, including the noise model, information-matched diffusion
schedule, and the lightweight conditioning mechanism used to modulate model
parameters as a function of the diffusion coordinate.

\subsection{Reverse diffusion noise model}
\label{app:diffusion_noise}

Let $\vEnc \in \mathbb{C}^{T \times D}$ denote the latent auxiliary ladder, where
$T$ indexes auxiliary tokens (including depth) and $D$ indexes complex channels.
Each $(t,d)$ entry is treated as an independent complex random variable with a
shared noise scale across its real and imaginary components.

Diffusion noise is sampled from an isotropic radial Laplace distribution on
$\mathbb{C} \simeq \mathbb{R}^2$,
\begin{equation}
p(\vNois_{t,d}) \propto
\exp\!\left(-\frac{|\vNois_{t,d}|}{\diffStd_{t,d}}\right),
\end{equation}
where $\diffStd_{t,d} > 0$ is a per-channel scale parameter. The radial component
follows a Gamma distribution with shape $k=2$ and scale $\diffStd_{t,d}$, yielding
\[
\mathbb{E}[|\vNois_{t,d}|] = 2\diffStd_{t,d},
\qquad
\mathbb{E}[|\vNois_{t,d}|^2] = 6\diffStd_{t,d}^2.
\]

During training, $\diffStd_{t,d}$ is estimated as a running average of the latent
channel magnitudes,
\begin{equation}
\diffStd_{t,d}
\;\leftarrow\;
\tfrac{1}{2}\,\mathbb{E}\!\left[\,|\vEnc_{t,d}|\,\right],
\end{equation}
where the expectation is taken over all training samples. This estimate is
computed without gradient tracking, accumulated over a single training epoch,
and then frozen. The resulting per-channel noise scales are used during sampling
and reverse diffusion, but do not participate in VAE training.

\subsection{Gaussian moment matching and variance calibration}
\label{app:moment_matching}

To relate the Laplace noise model to standard diffusion formulations, we approximate
each complex channel by an isotropic Gaussian with matched second moment. For a
radial Laplace distribution in two dimensions with scale $\diffStd$, the expected
squared norm is $\mathbb{E}[\|\vNois\|_2^2] = 6\diffStd^2$. Matching this to a
complex Gaussian with variance $\omega^2$ per real component yields
\begin{equation}
\mathbb{E}[\|\vNois\|_2^2] = 2\,\omega^2
\quad\Rightarrow\quad
\omega^2 = 3\diffStd^2.
\end{equation}

This moment-matched Gaussian approximation is used solely to calibrate the
diffusion schedule and does not alter the underlying Laplace noise model used
during sampling.

\subsection{Information-matched diffusion schedule}
\label{app:diffusion_schedule}

A scalar diffusion coordinate $\phi \in [\phi_{\min}, \phi_{\max}] \subset (0,1)$
is sampled uniformly and shared across the batch. Sharing $\phi$ is a memory
optimization that substantially reduces activation overhead while still covering
the full diffusion trajectory over training.

The forward diffusion process is defined as
\begin{equation}
\vEnc_\phi = c_\phi\,\vEnc + s_\phi\,\vNois,
\qquad
c_\phi^2 + s_\phi^2 = 1,
\end{equation}
where $\vNois$ is drawn from the Laplace distribution above. To define an
information-aware schedule, we analyze the process under the Gaussian
approximation from Appendix~\ref{app:moment_matching}.

Let $\exp(\vSigma)$ denote the learned decoder noise scale associated with each
latent channel. Under the Gaussian approximation, the effective signal-to-noise
ratio is
\begin{equation}
R = \frac{\omega^2}{\exp(2\vSigma)}
= 3\left(\frac{\diffStd}{\exp(\vSigma)}\right)^2.
\end{equation}

We choose mixing coefficients
\begin{equation}
s_\phi^2 = \frac{(1+R)^\phi - 1}{R},
\qquad
c_\phi^2 = 1 - s_\phi^2,
\end{equation}
so that the expected information gained when moving from pure noise
($\phi = 1$) to an intermediate state $\vEnc_\phi$ is approximately proportional
to $1-\phi$. In particular, $\phi=0$ corresponds to the pure signal distribution,
while $\phi=1$ corresponds to the maximally corrupted prior. This yields a
diffusion schedule in which $\phi$ is approximately linear in remaining
information content.

\subsection{Quadratic conditioning on diffusion coordinate}
\label{app:quadratic_conditioning}

Several model components are conditioned on the diffusion coordinate $\phi$ using
a lightweight quadratic interpolation scheme. For any parameter tensor $\Theta$
to be conditioned, three control tensors
$\{\Theta^{(0)}, \Theta^{(1)}, \Theta^{(2)}\}$ are learned, corresponding to
$\phi=0$, $\phi=\tfrac{1}{2}$, and $\phi=1$. The interpolated parameter is given by
\[
\Theta(\phi)
= B_0(\phi)\,\Theta^{(0)}
+ B_1(\phi)\,\Theta^{(1)}
+ B_2(\phi)\,\Theta^{(2)},
\]
with basis functions $B_0 = 2(0.5-\phi)(1-\phi)$,
$B_1 = 4\phi(1-\phi)$,
and $B_2 = 2\phi(\phi-0.5)$.
This scheme provides smooth, degree-2 modulation of parameters as a function of
$\phi$ without introducing additional networks or conditioning paths.

In practice, quadratic conditioning is applied to additive biases in complex RMS
normalization, post-gating MLP biases, learned RoPE3D phase offsets, and per-head
output scaling after attention.

\section{Additional Experimental Results}
\label{app:additional_results}

\subsection{Extended compression sweep results}
\label{app:compression_sensitivity}

This subsection reports robustness checks for auxiliary-ladder compression
hyperparameters, including the target nonzero-element (nnz) budget and the
compression schedule length. Results for the compression-strength sweep and the
schedule-length sweep are reported in
Tables~\ref{tab:cfactor} and~\ref{tab:nnzsteps}, respectively.

Table~\ref{tab:cfactor} varies the target nnz factor $C$ while holding the nnz
schedule length fixed at $160$k steps. Moderate compression achieves the best
overall reconstruction quality, as reflected by the lowest total VAE loss and
balanced lattice and Fourier reconstruction errors. More aggressive compression
reduces latent magnitude but degrades reconstruction, while weaker compression
yields no additional benefit.

Table~\ref{tab:nnzsteps} varies the length of the cosine nnz schedule for a fixed
target compression factor $C=8/12$. Performance varies smoothly with schedule
length, with an intermediate schedule providing the best trade-off between
reconstruction accuracy and latent suppression. Across both sweeps, the full VAE
objective $\lVae$ serves as the primary selection criterion, as it jointly
accounts for lattice reconstruction, Fourier reconstruction, species prediction,
and latent regularization.

\begin{table}[h]
    \centering
    \caption{\textbf{Compression strength sweep} (vary target nnz factor $C$; baseline nnz schedule length $160$k steps).
    ``C factor'' sets the final target nnz as $6(1+n_{\mathrm{elements}}+\text{bpd}^3)\,C$.
    Best (lowest) values per column are bolded.}
    \label{tab:cfactor}
    \begin{tabular}{c||c|c|c|c|c}
    \toprule
    C factor & $\lVae$ & $\lFour$ & $\lLat$ & $\lMu$ & $\lElt$ \\
    \midrule
    7/12 & 0.268 & 0.235 & 0.073 & \textbf{0.034} & $\bm{<}$0.001 \\
    8/12 & \textbf{0.221} & \textbf{0.209} & \textbf{0.034} & 0.043 & 0.001 \\
    9/12 & 0.254 & 0.223 & 0.068 & 0.041 & 0.001 \\
    \bottomrule
    \end{tabular}
\end{table}

\begin{table}[h]
    \centering
    \footnotesize
    \caption{\textbf{Compression schedule length sweep} (vary ``nnz steps''; baseline target nnz factor $C=8/12$).
    ``nnz steps'' is the number of optimization steps over which the cosine schedule reaches the final nnz target.
    Best (lowest) values per column are bolded.}
    \label{tab:nnzsteps}
    \begin{tabular}{c||c|c|c|c|c}
    \toprule
    nnz steps & $\lVae$ & $\lFour$ & $\lLat$ & $\lMu$ & $\lElt$ \\
    \midrule
    120k & 0.241 & 0.205 & 0.071 & \textbf{0.036} & 0.003 \\
    160k & \textbf{0.221} & 0.209 & \textbf{0.034} & 0.043 & 0.001 \\
    200k & 0.230 & \textbf{0.198} & 0.068 & 0.040 & 0.001 \\
    \bottomrule
    \end{tabular}
\end{table}

\subsection{Reverse diffusion fine-tuning robustness}
\label{app:rd_robustness}

This subsection reports robustness checks for reverse diffusion fine-tuning across
a sweep of learning rates, evaluated for both the baseline architecture and a
variant with $N_{\mathrm{head}}=10$. Results are summarized in
Tables~\ref{tab:rd_ft_sweep_baseline} and~\ref{tab:rd_ft_sweep_nh10}, respectively.

Metrics follow the main text: noise RMSE in the signal-dominant regime (low-$\phi$),
mean noise RMSE across the diffusion trajectory, and one-shot signal reconstruction
RMSE in the signal-dominant regime (diagnostic). The low-$\phi$ bin corresponds to
$\phi \in [0.01, 0.2]$, where inputs are largely signal-dominated and denoising is
most challenging.

Across both architectures, performance varies smoothly over more than an order of
magnitude in learning rate, with no sharp instabilities observed. The
$N_{\mathrm{head}}=10$ architecture consistently achieves lower noise RMSE than the
baseline, both in the signal-dominant regime and on average across the diffusion
schedule. In all cases, signal reconstruction error in the low-$\phi$ regime
remains below the decoder noise scale, indicating that reconstructed signals stay
within the operating range of the decoder.

\begin{table}[t]
\centering
\caption{\textbf{Meta-stable fine-tuning LR sweep (Baseline model).}
All values are RMSE. ``low-$\phi$'' corresponds to $\phi \in [0.01,0.2]$.}
\label{tab:rd_ft_sweep_baseline}
\begin{tabular}{c||c|c|c}
\toprule
lr & low-$\phi$ noise & mean noise & low-$\phi$ sig. \\
\midrule
$5\times10^{-5}$ & 5.464 & 3.027 & 0.634 \\
$2\times10^{-5}$ & 5.419 & \textbf{3.018} & \textbf{0.633} \\
$1\times10^{-5}$ & \textbf{5.417} & 3.025 & 0.635 \\
$5\times10^{-6}$ & 5.427 & 3.035 & 0.636 \\
$2\times10^{-6}$ & 5.452 & 3.052 & 0.639 \\
\bottomrule
\end{tabular}
\end{table}

\begin{table}[t]
\centering
\caption{\textbf{Meta-stable fine-tuning LR sweep ($N_{\mathrm{head}}=10$ model).}
All values are RMSE. ``low-$\phi$'' corresponds to $\phi \in [0.01,0.2]$.}
\label{tab:rd_ft_sweep_nh10}
\begin{tabular}{c||c|c|c}
\toprule
lr & low-$\phi$ noise & mean noise & low-$\phi$ sig. \\
\midrule
$5\times10^{-5}$ & 4.700 & 2.665 & 0.604 \\
$2\times10^{-5}$ & 4.674 & \textbf{2.661} & \textbf{0.604} \\
$1\times10^{-5}$ & 4.675 & 2.666 & 0.605 \\
$5\times10^{-6}$ & 4.683 & 2.674 & 0.607 \\
$2\times10^{-6}$ & 4.702 & 2.686 & 0.609 \\
\bottomrule
\end{tabular}
\end{table}

\section{Limitations}
Our approach relies on truncated Fourier representations and rational-grid snapping, which introduce controlled discretization error and impose resolution-dependent recoverability limits. Increasing Fourier resolution improves fidelity but incurs additional computational cost.

Coordinate recovery from Fourier coefficients is performed via a multistage, non-differentiable procedure. While this decouples training from decoding, it may limit robustness for extremely dense unit cells.

Finally, unconditional latent diffusion is observed to amplify dataset skew toward the small-cell regime. Although this does not affect reconstruction or denoising capability, it limits unconditional generation diversity and motivates future work on conditional generation, size-aware priors, or guided diffusion.

\end{document}